\documentclass[10pt,journal,compsoc]{IEEEtran}
\usepackage{graphicx}
\usepackage{pgfplots}
\usepackage{subcaption}
\usepackage{amssymb}
\usepackage{rotating}
\usepackage{tikz}
\usepackage{pgfplotstable}
\usepackage{color}
\usetikzlibrary{plotmarks}
\usetikzlibrary{pgfplots.groupplots} 
\usepackage{placeins}
\usepackage[utf8]{inputenc}
\usepackage{forest}
\usetikzlibrary{trees,positioning,shapes,shadows,arrows.meta}
\usepackage[EULERGREEK]{sansmath}

\usepackage{multirow}
\usepackage[a4paper,margin=2cm]{geometry}
\usepackage{booktabs,makecell,tabularx}

\usepackage{array,ragged2e,longtable}
\usetikzlibrary{positioning,plotmarks,pgfplots.groupplots}

\newcolumntype{C}[1]{>{\centering\arraybackslash}p{#1}}
\newcolumntype{L}{>{\raggedright\arraybackslash}X}
\usepackage{titlesec}
\usepackage{colortbl}

\usepackage{siunitx}
\usepackage{etoolbox}
\newrobustcmd{\B}{\bfseries}
\pgfplotsset{compat=1.18}
\usetikzlibrary{shapes.geometric}
\graphicspath{ {./images/} }
\newcommand{\subfour}[1]{\vspace*{3mm}{\noindent\bf #1}}

\titlespacing\section{0pt}{12pt plus 4pt minus 2pt}{0pt plus 2pt minus 2pt}
\titlespacing\subsection{0pt}{12pt plus 4pt minus 2pt}{0pt plus 2pt minus 2pt}
\titlespacing\subsubsection{0pt}{12pt plus 4pt minus 2pt}{0pt plus 2pt minus 2pt}

\usepackage{amsmath}

\DeclareMathOperator*{\argmin}{arg\,min}

\ifCLASSOPTIONcompsoc
  \usepackage[nocompress]{cite}
\else
  \usepackage{cite}
\fi
\ifCLASSINFOpdf
\else
\fi

\begin{document}
\title{Handling Out-of-Distribution Data: A Survey}

\author{Lakpa~Tamang,
        Mohamed~Reda~Bouadjenek,
        Richard~Dazeley,~and~Sunil~Aryal
\thanks{School of Information Technology, Deakin University, Geelong Waurn Ponds Campus, Australia.
E-mail: l.tamang@research.deakin.edu.au, \{reda.bouadjenek, richard.dazeley, sunil.aryal\}@deakin.edu.au
}
\thanks{Manuscript received Xxx XX, 2025; revised Xxxx XX, 2025.}}

\markboth{Journal of \LaTeX\ Class Files,~Vol.~14, No.~8, August~2015}%
{Shell \MakeLowercase{\textit{et al.}}: Bare Demo of IEEEtran.cls for Computer Society Journals}

\IEEEtitleabstractindextext{%
\begin{abstract}
In the field of Machine Learning (ML) and data-driven applications, one of the significant challenge is the change in data distribution between the training and deployment stages, commonly known as distribution shift. This paper outlines different mechanisms for handling two main types of distribution shifts: (i) \textbf{Covariate shift: } where the value of features or covariates change between train and test data, and (ii) \textbf{Concept/Semantic-shift: } where model experiences shift in the concept learned during training due to emergence of novel classes in the test phase. We sum up our contributions in three folds. First, we formalize distribution shifts, recite on how the conventional method fails to handle them adequately and urge for a model that can simultaneously perform better in all types of distribution shifts. Second, we discuss why handling distribution shifts is important and provide an extensive review of the methods and techniques that have been developed to detect, measure, and mitigate the effects of these shifts. Third, we discuss the current state of distribution shift handling mechanisms and propose future research directions in this area. Overall, we provide a retrospective synopsis of the literature in the distribution shift, focusing on OOD data that had been overlooked in the existing surveys.
\end{abstract}

\begin{IEEEkeywords}
Data Distribution Shift, Out-of-Distribution, Covariate Shift, Concept Shift
\end{IEEEkeywords}}

\maketitle

\IEEEdisplaynontitleabstractindextext

\IEEEpeerreviewmaketitle

\IEEEraisesectionheading{\section{Introduction}\label{sec:introduction}}

\IEEEPARstart{E}{xisting} Machine Learning (ML) techniques, particularly Deep Neural Networks (DNNs), have shown unprecedented success across domains, such as computer vision, natural language processing, and recommendation systems \cite{sarker_deep_2021}. These models tend to exploit subtle statistical correlations present in the training distribution, yielding impressive results under the \textit{i.i.d} (independently and identically distributed) hypothesis. However, despite their prowess under controlled experimental conditions, there is growing empirical evidence highlighting their vulnerabilities to real-world data distribution shifts. These shifts may often surface in relation to several confining factors, such as sample selection biases, non-stationary environments, and other inherent peculiarities of data generation mechanisms \cite{moreno-torresUnifyingViewDataset2012}. As demonstrated by the adversarial examples in \cite{szegedyIntriguingPropertiesNeural2014}, even subtle changes in the data distribution can have a significant impact on the performance of advanced classifiers. Therefore, it is imperative to understand and address these vulnerabilities, especially for systems that perform safety-critical or high-impact operations such as medical diagnosis and autonomous vehicles.
\par
\begin{figure}[ht!]
    \centering
    \includegraphics[width=0.5\textwidth]{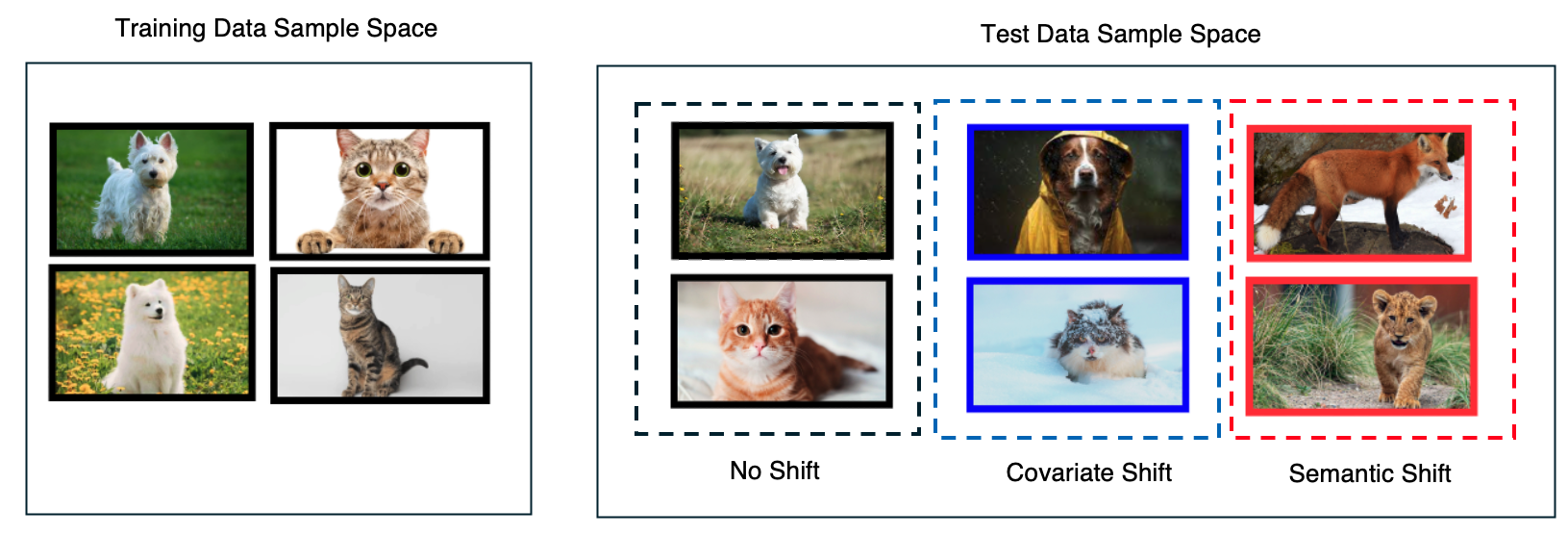}
    \caption{Schematic illustration of data distribution shift between training and testing data sample spaces. In general, a model is trained on a limited knowledge of real-world samples, but it can encounter a whole set of similar or differently distributed inputs either in feature or semantic spaces when subjected to testing in the wild.}
    \label{fig:introplot}
\end{figure}

 The change in data distribution can hamper the model's accuracy, making its result unreliable to adapt to. One of several factors instigating these changes is the bias introduced by the experimental design due to the inimitability of the inconsistent testing conditions during training \cite{quinonero-candelaDatasetShiftMachine2009}. In other words, the knowledge that the model has learned from the training data may be sampled under conditions different from the ones it will encounter during actual testing. For graphical illustration, we refer to Fig. \ref{fig:introplot} where the large corpus of similarly and distinctly distributed test data (in feature, i.e., \textit{covariates} space or semantic space) are put against a limited number of instances from the training sample space. While it is impractical to address all real-world test cases, it is crucial that the model be capable of handling such variations without compromising the relevance of the deployed model. Moreover, to make informed decisions about when and how to update the model, complying to the changes in the distribution of data that affect the model's performance is very important.

\par It is worthy to note that best practices for detecting shifts in high-dimensional real-world data have not yet been definitively established \cite{alemiUncertaintyVariationalInformation2018}. Regardless, numerous studies have been proposed with the primary objective of addressing the changes in data distribution by adapting and generalising to distributionally shifted samples or rejecting them entirely. In practice, the data distribution can be shifted in one of two ways: in feature space (\textit{covariate shift}) or in label space (\textit{semantic} or \textit{concept shift}). Although numerous review papers have been released discussing the strategies for effectively addressing individual shifts, our review is the inaugural one to recognize these shifts as a collective issue.  Rather than constraining the work into specifics of a single type of distributional shift \cite{liu2021towards}, \cite{yang2021generalized}, we aim to emphasize the topic into a broader spectrum of research focusing on how each type of distributional shift is handled to make a model reliable in practice. In this paper, we present a comprehensive survey of the different modelling strategies dealing with data distribution change. Particularly, we aim to provide a comprehensive and nuanced understanding of the topic to the researchers by focusing on the retrospective overview of different methodologies centralized around handling data distribution shift problems as a generalized entity. In summary, through this paper, we offer three distinct contributions to the academic community in this domain of research:
\begin{enumerate}
  \item \textbf{Formalization of Shifts} We formalize the  prevalent types of shift and advocate that modelling strategies should account for both generalization and rejection when subjected to data distribution shifts. In accordance with this theory, we link existing topics by examining them independently to help the research community understand the practical objective of building ML models that are robust enough to handle both types of distribution shifts.
  \item \textbf{Comprehensive Review of Modelling Strategies} In light of the presence of other literature reviews in the same field \cite{liu2021towards}, \cite{yang2021generalized}, \cite{geng2020recent}, \cite{zhou2022domain} our work stands out by providing a comprehensive overview of key discoveries in the field of related topics, specifically centred around shift handling objective.
  \item \textbf{Future Research Directions} We aim to provide readers with a deeper understanding of the current challenges and opportunities in this topic while also highlighting potential future research directions.
\end{enumerate}

\begin{figure}
    \centering
    \includegraphics[width=0.5\textwidth]{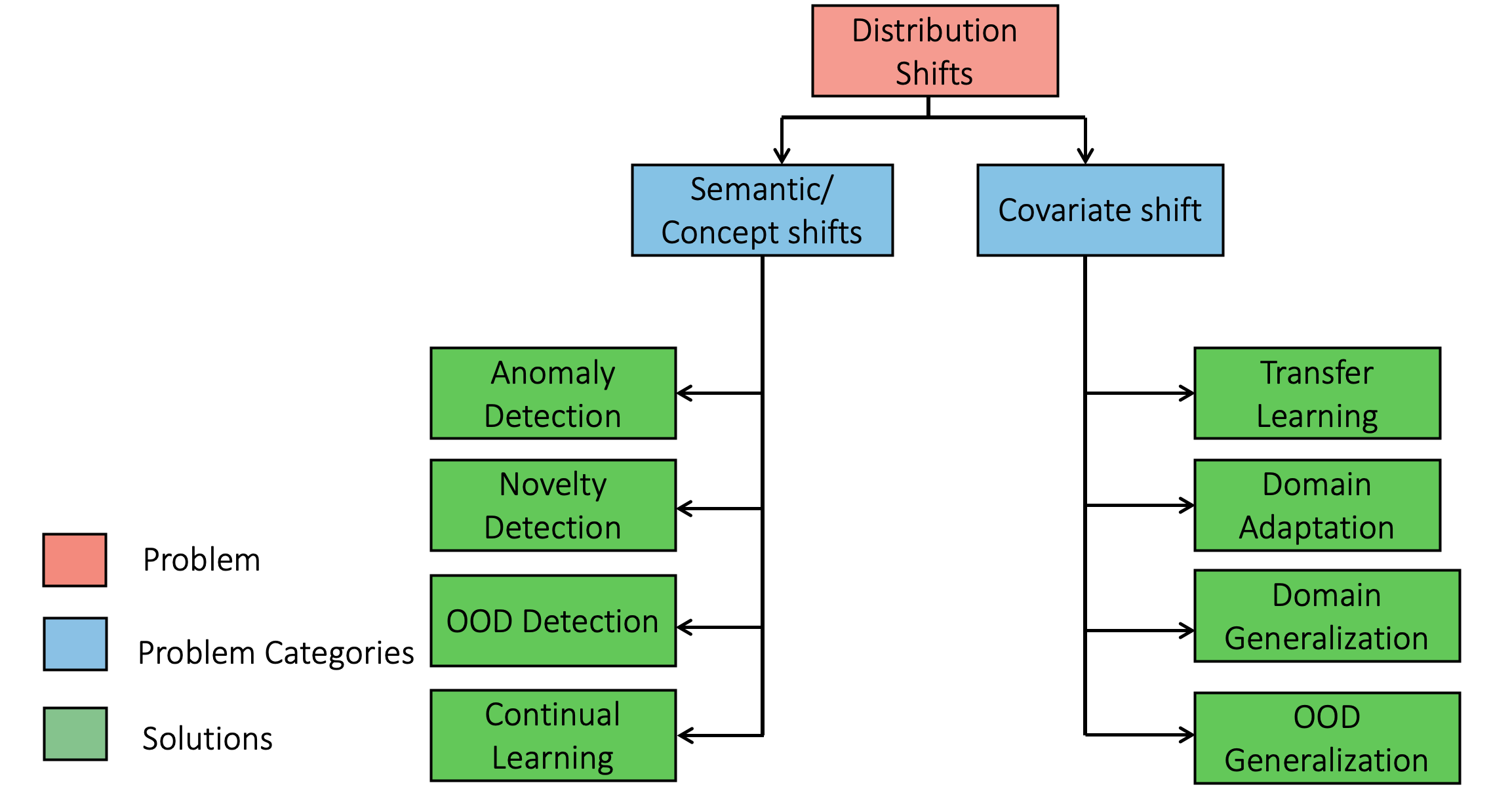}
    \caption{Taxonomy of different methodologies for handling data distribution shift.}
    \label{fig:taxonomy_one}
\end{figure}

\section{Background}
\subsection{Data Distribution}
\noindent Data distribution refers to the arrangement of data values in a dataset and provides insights into patterns, characteristics, and relationships within the data \cite{pivk2005plots}. It is crucial to understand data distribution that spans over a wide range of topics, including statistics, machine learning, and data analysis, in order to establish one of many objectives such as discovering trends, identifying outliers, and making informed decisions. The distribution of data obtained from a sample is vital in understanding how to analyze it, as it provides a parameterized mathematical function that can be used to calculate the probability for any individual observation from the sample space. In the field of statistics, different kinds of data distributions exist, such as normal \cite{bryc2012normal}, uniform \cite{kuipers2012uniform}, skewed \cite{arellano2005fundamental}, and bimodal \cite{murphy1964one}, where each has their unique characteristics associated with the sample. In Machine learning and Deep Learning, probability distributions are considered to model real-world data and make predictions \cite{murphy2012machine}, \cite{bishop2006pattern}. A probability distribution \cite{ramberg1979probability} is a mathematical function that describes the likelihood of different outcomes for a random variable. It allows for the quantification of uncertainty and the making of predictions based on past data. These algorithms often involve estimating probability distributions from sample data and using them to generalize to new examples.
\subsection{Why Data Distribution Changes?}
\noindent There is an assumption that the distributions specified by unconditional or conditional models are static, remaining unchanged from the time they are learned to the time they are used \cite{quinonero-candelaDatasetShiftMachine2009}. However, if this assumption is not true and the distributions undergo some kind of change, then we must account for this change or at least the possibility of it. This requires examining the reasons why such a shift may occur. There are several reasons why an ML model might exhibit a data distribution shift.

\subfour{Bias During Sample Selection: }The concept of sample selection bias refers to a fault in the process of collecting or labelling data that leads to the uneven distribution of training examples. This results from the fact that the training examples were obtained through a biased method, which means they may not accurately reflect the environment where the classifier will be used. {During the sampling process of the training data, the data points $x^{te}_i$ may not precisely represent the actual testing distribution $P^{te}(X,Y)$}. For instance, while generating a handwritten dataset, one may get rid of an entirely obscure character, although it may hold true that some characters are more likely to be written in an unclear manner.

\subfour{Deployment Environment Changes: } It is often true that data remains non-stationary to time and space change \cite{alaiz-rodriguezImprovingClassificationChanges2009}. Environments are dynamic in general, and sometimes the difficulties of matching the learning scenario (with training data) to the real-world use (test data) are constrained by these changes. Such scenarios make it challenging to develop an understanding of the appropriateness of a particular model in the circumstance of these environmental changes, thus the prevalence of the shift. For example, a commonly observed issue with ML models trained to predict electricity demand based on historical data of usage time, temperature, and humidity is their failure when non-stationary changes such as climate and the adoption of renewable energy sources occur.

\subfour{Change in the Domain: } Occasionally, a new sample might be collected from a different domain to represent the same category. In this regard, various domains may use various terms to refer to the same entity. The changes in the domain are characterized by the fact that the measurement system or the description technique of the feature of a dataset is changed.  For example, in a computer vision scenario, the change in visual concepts such as illumination conditions, image resolution or background of $x^{te}_i$, relative to $x^{tr}_i$ might contribute to the domain shift \cite{tommasiLearningRootsVisual2016}.

\subfour{Existence of Uncategorized Instances: } The closed space assumption of traditional ML algorithms certainly doesn't hold true in the open world where unseen situations can emerge unexpectedly \cite{gengRecentAdvancesOpen2021}. It is an inherent fact that the test set may contain some classes that are not present in the training set \cite{xiaSpatialLocationConstraint2021}. Apparently, the model may experience a shift in the semantics of the representation it has learned from the training set as a result of the appearance of these unseen instances. For example, if a binary classifier trained with categories of dog and cat suddenly sees a fox, whilst the covariates of dog and fox might have some correlation, they represent entirely different semantics.


\section{Formalizing Distribution Shifts}
\noindent In this section, we will formalize  different distribution shifts by adhering to the official definition of the topic presented in \cite{moreno-torresUnifyingViewDataset2012} and building upon it in terms of addressing the problem. Abbreviations used in the paper are enlisted in Table. \ref{tab:abbreviations}.
\begin{table}[]
    \centering
    \caption{{List of abbreviations used throughout the paper.}}
    \begin{tabular}{l l}
    \toprule
        \textbf{Abbreviations} & \textbf{Full forms} \\
        \midrule
        \midrule
        FID & Frechet Inception Distance \\
        LPIPS & Learned Perceptual Image Patch Similarity \\
        RMSE & Root Mean Squared Error \\
        MAE & Mean Absolute Error \\
        TPR & True Positive Rate \\
        FPR & False Positive Rate \\
        TNR & True Negative Rate \\
        KID & Kernel Inception Distance \\
        IoU & Intersection of Union \\
        FDR & False Detection Rate \\
        AUC & Area Under Curve \\
        NMI & Normalized Mutual Index \\
        AP & Average Precision \\
        AUROC & \makecell{Area Under Receiver Operating Characteristic \\ Curve} \\
        AUPR & Area Under Precision Recall Curve \\
        FPR & False Positive Rate \\
        OSCR & Open Set Classification Rate \\
        BWF & Backward Forgetting \\
        FWT & Forward Transfer \\
        BWT & Backward Transfer \\
        MCR & Mean Class Recall \\
        \bottomrule
    \end{tabular}
    \label{tab:abbreviations}
\end{table}
\subsection{Preliminaries and Definitions}
\noindent Let $\left\{(x_1^{tr}, y_1^{tr}), (x_2^{tr}, y_2^{tr}), ..., (x_n^{tr}, y_n^{tr})\right\}$ be the labelled training data sampled from distribution $\mathcal{D}^{tr}(\textit{X}, \textit{Y})$, where $x_i^{tr}\in \textit{X}$,~and $y_i^{tr}\in \textit{Y}$ represent the $i^{th}$ sample and the associated label, respectively. Similarly, let $\left\{(x_1^{te}, y_1^{te}), (x_2^{te}, y_2^{te}), ..., (x_m^{te}, y_m^{te})\right\}$ be the test data sampled from a test distribution $\mathcal{D}^{te}(\textit{X},\textit{Y})$. 
Let, $P^{tr}(X)$, $P^{tr}(Y)$ be the marginal distributions and let $P^{tr}(Y|X)$, and $P^{te}(Y|X)$ be the conditional distributions for the training and test data respectively. Based on this, we define the following:

\subfour{Definition 1: } (No-Shift) 
\textit{The test data is said to be not shifted (in other words in \textbf{in-distribution (ID)} with the training data) when ${P}^{tr}(\textit{X}, \textit{Y})={P}^{te}(\textit{X}, \textit{Y})$. In this scenario, the statistical properties of the data (both the marginal and conditional distribution of the input variables) are assumed to be same between the training and testing phases. i.e., $P^{tr}(X) = P^{te}(X)$ and $P^{tr}(Y|X) = P^{te}(Y|X)$}.

If $L(f(x), y)$ is the loss function for some particular pair of inputs $X$, and outputs $Y$, we define following:

\subfour{Definition 1.1 (True Risk): } \textit{is mathematically defined as}:

\begin{equation}
    R_{true}(f) = \mathop{\mathbb{E}}_{p_{true}}[L(f(X), Y)]
\end{equation}
where $p_{true}$ is the true distribution over the $x$, and $y$ which is unknown.

\subfour{Definition 1.2 (Empirical Risk): }\textit{is mathematically defined as}:

\begin{equation}
    R_{emp}(f) = \mathop{\mathbb{E}}_{p_{emp}}[L(f(X), Y)] = \frac{1}{n}\sum_{i=1}^n L(f(X_i), Y_i)
\end{equation}
where $p_{emp}$ is the sampled distribution which consists of limited number of samples quantitatively smaller than $p_{true}$ and can lie in different regions of sample space of $p_{true}$. Under no-shift condition, which is a fundamental ground of i.i.d hypothesis, it is generally assumed that the empirical risk minimization (ERM) \cite{vapnik1998statistical} leads to consistent generalization. Since $\mathcal{D}_{tr}$ is the representative of $\mathcal{D}_{te}$, the model trained using ERM will perform similarly on the test data. With large enough training samples, $p_{emp}$ approximates $p_{true}$ well and empirical risk converges to the true risk i.e., $R_{emp}(f)\rightarrow R_{true}(f)$. This renders that for any $f \in \mathcal{F}$, minimizing the empirical risk minimization performs similarly to true risk minimization leading to optimal prediction performance.

\begin{equation}
    \mathop{\argmin}_{f}R_{emp}(f) \approx \mathop{\argmin}_{f}R_{true}(f)
\end{equation}

\par Conversely, the test data is said to be Out-Of-Distribution (OOD) when $\mathcal{D}^{tr}(\textit{X}, \textit{Y})\not= \mathcal{D}^{te}(\textit{X}, \textit{Y})$. In this scenario, the model trained with ERM will perform poorly as $R_{emp}(f)$ does not accurately reflect $R_{true}(f)$, and $p_{true} \not= p_{emp}$.  Under the OOD framework, we define two independent distribution shifts as follows:

\subfour{Definition 2: } (OOD with Covariate Shift)
\textit{The test data is in OOD with covariate shift, when it is subjected to change of distribution in feature space i.e., ${P}^{tr}(X)\not={P}^{te}(X)$, but, the conditional distribution of the target given the input remains unchanged i.e., ${P}^{tr}(Y|X)={P}^{te}(Y|X)$}. 
\par {As can be seen from Fig. \ref{fig:introplot}, the training data contains images of dog sitting front of a grass, whereas in the test space (highlighted in blue) the dog appears to be wearing a raincoat in front of a dark background. Here, although the conditional distributions (labels of training and test) should remain same, the features representing the respective labels are distinct. In such cases, based on the inductive biases, the learning algorithm may abruptly fail to correctly classify the test samples.}

\subfour{Definition 3: } {(OOD with Semantic Shift)
\textit{The test data is said to be in semantic shift with the training data when their relationship between the input and the target variables change i.e., $P^{tr}(Y|X)\not= P^{te}(Y|X)$. This type of shift can occur regardless of whether the margin distributions of the input variables change i.e., $P^{tr}(X)\not=P^{te}(X)$} or remains the same $P^{tr}(X)=P^{te}(X)$}. \par {In Fig. \ref{fig:introplot}, although the test samples highlighted in red share visual similarity to the samples of $P^{tr}(X)$ (fox might look like a dog, and a cub might look like a cat), the conditional distribution of these samples entirely differ. In this example, the features $P^{tr}(X)$ and $=P^{te}(X)$ share some similarity, while the semantic shift might occur due to entirely different features as well. For instance, a test sample being any non-lookalike object to any of the training set.}

{\subsection{Real-world Shift Dynamics}}
\noindent In practice, during the production phase, machine learning models are exposed to test data that does not strictly adhere to the training distribution. In fact, the test data can be shifted into one of the two prevalent shifts: covariate and semantic whose distributions are denoted as $\mathcal{D}^{te}_C$, and $\mathcal{D}^{te}_S$ respectively. Therefore, mathematically we can say:  
\begin{equation} \label{eq:1}
    \{ \mathcal{D}^{te}_C, \mathcal{D}^{te}_S\} \subset \mathcal{D}^{te}(\textit{X}, \textit{Y})
\end{equation}

Now that the i.i.d assumption is routinely violated, ERM does not account for the changes in distribution, leading to suboptimal generalization and prediction performance. Intuitively, when a data is drawn from the sample space of test dataset, then it can experience either experience no shift or be shifted in covariate space or semantic space. In the face of distribution shifts, it is highly important for machine learning models to move beyond the ERM framework to handle the OOD inputs. For instance, for a test sample $x_m^{te}$ drawn from $\mathcal{D}^{te}$, the objective should be minimizing the OOD risk $R_{OOD}$, which equates to minimizing risk for covariate, $R_{cov}$ and semantic shift, $R_{sem}$ simultaneously. 

\begin{equation}
    \mathop{\argmin}_{f}R_{OOD}(f)=\mathop{\argmin}_{f}R_{cov}(f) + \mathop{\argmin}_{f}R_{sem}(f)
\end{equation}

\section{Distribution Shifts Mitigation Strategies}
\noindent In this section, we discuss different approaches employed in the literature to handle different types of distribution shifts. Specifically, we review the modelling strategies in two perspectives; the first one being related to the feature space shift (covariate shift), and the other one associated with the shift in concepts (semantic shift). For each category, we review several approaches by focusing on their corresponding representative works in terms of types of problem domains such as computer vision and natural language processing. Moreover, a global picture of how these approaches are linked to solving a common problem of data distribution shift is discussed.
\begin{table*}[htbp]
\centering
\small
\renewcommand{\arraystretch}{1.35}
\rowcolors{2}{gray!5}{white}
\caption{{Systematic Comparison of Different Types of Shift Mitigation Strategies}}
\label{table:comparison}

\begin{subtable}{\textwidth}
\centering
\caption{{Covariate Shift}}
\label{table:covariate_comp}
\resizebox{\textwidth}{!}{
\begin{tabular}{>{\bfseries}l p{1.8in} p{1.8in} p{1.8in}}
\rowcolor{gray!20}
\textbf{Criterion} & \textbf{Transfer Learning} & \textbf{Domain Adaptation} & \textbf{Domain Generalization} \\
\toprule
Distribution Assumption & $P^{te}(X)$ is different but related to $P^{tr}(X)$ & $P^{te}(X)$ is shifted but has overlap with $P^{tr}(X)$ & $P^{te}(X)$ is entirely unknown \\
Data Access & Access to labeled $P^{te}(X)$ & Access to unlabeled $P^{te}(X)$ & No access to $P^{te}(X)$ \\
Learning Method & Pretraining and fine-tuning & Source training and adaptation & Contrastive learning, IRM, Feature disentanglement learning \\
Testing Strategy & Evaluate on fine-tuned target & Compare performance before and after adaptation & Measure zero-shot generalization \\
Model Selection & Best pre-trained model for fine-tuning & Best adaptation method per domain pair & Best generalization across domains \\
Regularization Level & Minimal & High (to prevent overfitting to source domains) & Very high (for domain-invariant learning) \\
Generalization Capability & Poor & Moderate & Strong \\
\bottomrule
\end{tabular}}
\end{subtable}

\vspace{0.7cm}

\begin{subtable}{\textwidth}
\centering
\caption{{Semantic Shift}}
\label{table:semantic_comp}
\resizebox{\textwidth}{!}{
\begin{tabular}{>{\bfseries}l p{1.6in} p{1.6in} p{1.6in} p{1.6in}}
\rowcolor{gray!20}
\textbf{Criterion} & \textbf{Anomaly Detection} & \textbf{OOD Detection} & \textbf{OSR} & \textbf{CL} \\
\toprule
Core Objective & Detecting rare, extreme outliers & Detect entirely different distributions & Balance ID classification and unknown rejections & Learning with evolving data over time \\
Training Data Assumption & Only ID & OOD data may be available for regularization & ID and OOD data & ID and new OOD classes evolved over time \\
Test Data Assumption & Rare anomalies that share some features with ID & Samples from entirely different dataset or domain & Samples from ID and OOD distributions & Samples from old ID and newly evolved OOD \\
Data Availability & Labeled ID, unlabeled anomalies & Labeled ID, unlabeled or uniformly labeled OOD & Labeled ID and unlabeled OOD & Labeled past ID, unlabeled emerging OOD \\
\bottomrule
\end{tabular}}
\end{subtable}
\end{table*}

\begin{figure*}[ht!]
    \centering
    \begin{subfigure}[b]{0.24\linewidth}
        \includegraphics[width=\linewidth]{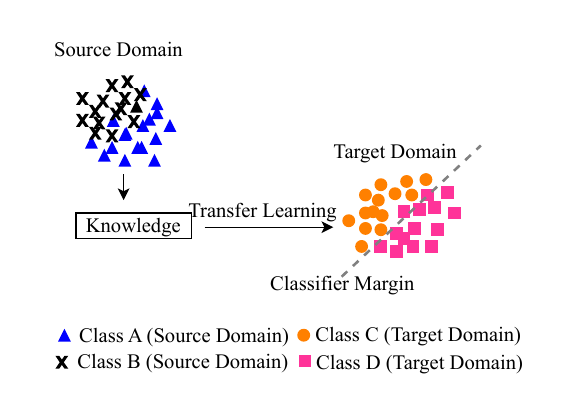}
        \caption{Transfer Learning}
    \end{subfigure}
    \hfill
    \begin{subfigure}[b]{0.24\linewidth}
        \includegraphics[width=\linewidth]{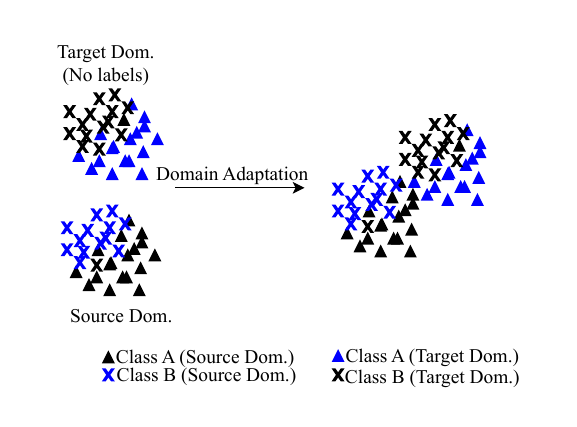}
        \caption{Domain Adaptation}
    \end{subfigure}
    \hfill
    \begin{subfigure}[b]{0.24\linewidth}
        \includegraphics[width=\linewidth]{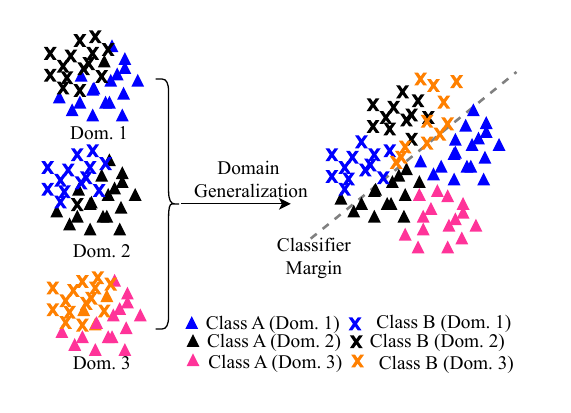}
        \caption{Domain Generalization}
    \end{subfigure}
    \hfill
    \begin{subfigure}[b]{0.24\linewidth}
        \includegraphics[width=\linewidth]{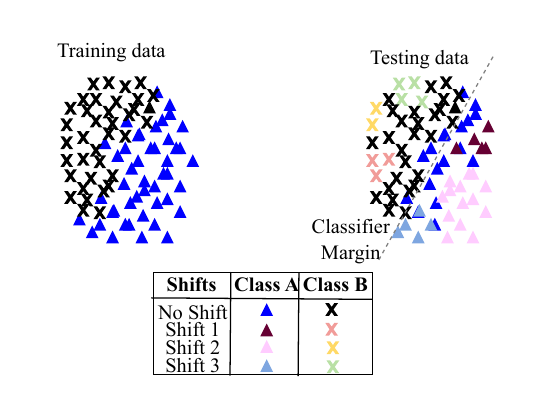}
        \caption{OOD Generalization}
    \end{subfigure}
    \caption{Schematic representation of different mitigation approaches for handling Covariate Shift.}
    \label{fig:covariate_schematic}
\end{figure*}

\begin{figure*}[ht!]
    \centering
    \begin{subfigure}[b]{0.24\linewidth}
        \includegraphics[width=\linewidth]{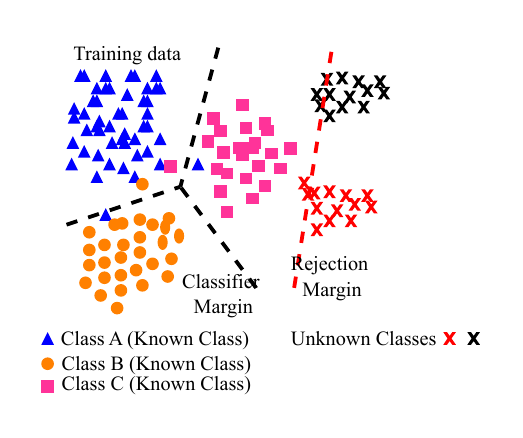}
        \caption{Open Set Recognition}
    \end{subfigure}
    \hfill
    \begin{subfigure}[b]{0.24\linewidth}
        \includegraphics[width=\linewidth]{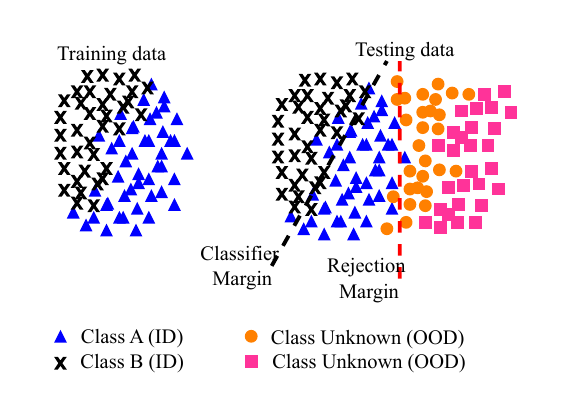}
        \caption{OOD Detection}
    \end{subfigure}
    \hfill
    \begin{subfigure}[b]{0.24\linewidth}
        \includegraphics[width=\linewidth]{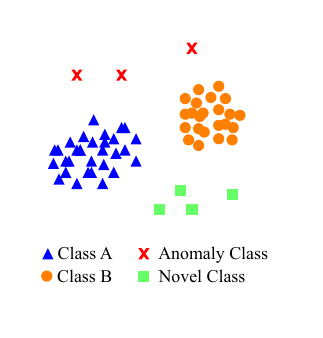}
        \caption{Anomaly/ Novelty Detection}
    \end{subfigure}
    \hfill
    \begin{subfigure}[b]{0.24\linewidth}
        \includegraphics[width=\linewidth]{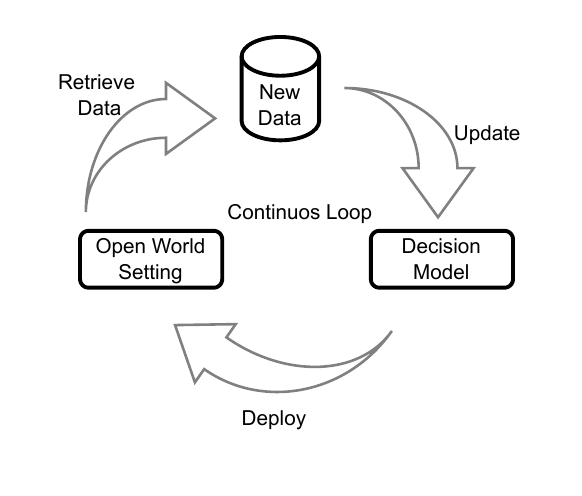}
        \caption{Continual Learning}
    \end{subfigure}
    \caption{Schematic representation of different mitigation approaches for handling Covariate Shift.}
    \label{fig:concept_schematic}
\end{figure*}

\subsection{Review Structure of the Paper}
\begin{figure*}[ht]
\centering
\begin{tikzpicture}
\begin{groupplot}[
    group style={
        group size=2 by 2,
        horizontal sep=2cm,
        vertical sep=2cm,
    },
    width=0.47\textwidth,
    height=0.27\textwidth,
    grid=both
]

\nextgroupplot[
        xmajorgrids = false,
    ymajorgrids = false,
    scatter/classes={
        a={mark=triangle,   color=blue},       
        b={mark=triangle, color=red},    
        c={mark=triangle, color=green!60!black}, 
        d={mark=triangle, color=orange},
        e={mark=triangle, color=violet},    
        f={mark=triangle, color=cyan!60!black},    
        g={mark=triangle, color=black},        
        h={mark=triangle, color=teal},     
        i={mark=triangle, color=magenta},         
        j={mark=triangle*, color=brown}           
    },     xlabel={AUROC},
    ylabel={AUCPR}, label style={font=\scriptsize},
    xmin=82, xmax=88,
    ymin=37, ymax=44.2,
    after end axis/.code={
    \node[anchor=north, font=\small] at (current axis.north) {(a)};
},
]

\addplot[scatter,only marks,scatter src=explicit symbolic]
table[meta=label] {
    x    y     label
    82.71 38.61  a
    86.92 44.06  b
    87.40 39.52  c
    85.33 39.39  d
    87.37 40.60  e
    86.12 42.50  f
    83.03 38.00  g
    87.32 42.03  h
    86.61 41.35  i
    87.63 42.28  j
};

\node at (axis cs: 82.71, 38.61) [anchor= south west] {\tiny DeepSAD \cite{ruff2019deep}};
\node at (axis cs: 86.92, 44.06) [anchor=north east] {\tiny MLP \cite{rosenblatt1958perceptron}};
\node at (axis cs: 87.40, 39.52) [anchor=north] {\tiny PReNet \cite{pang2023deep}};
\node at (axis cs: 85.33, 39.39) [anchor=south west] {\tiny SVM \cite{li2003improving}};
\node at (axis cs: 87.37, 40.60) [anchor=north] {\tiny REPEN \cite{pang2018learning}};
\node at (axis cs: 86.12, 42.50) [anchor=north west] {\tiny XGB \cite{chen2016xgboost}};
\node at (axis cs: 83.03, 38.00) [anchor=north west] {\tiny DevNet \cite{pang2019deep}};
\node at (axis cs: 87.32, 42.03) [anchor=north] {\tiny XGBOD \cite{zhao2018xgbod}};
\node at (axis cs: 86.61, 41.35) [anchor=north west] {\tiny LGB \cite{ke2017lightgbm}};
\node at (axis cs: 87.63, 42.28) [anchor=south] {\tiny CatB \cite{prokhorenkova2018catboost}};

\nextgroupplot[
        xmajorgrids = false,
    ymajorgrids = false,
    scatter/classes={
        a={mark=triangle,   color=blue},       
        b={mark=triangle, color=red},    
        c={mark=triangle, color=green!60!black}, 
        d={mark=triangle, color=orange},
        e={mark=triangle, color=violet},    
        f={mark=triangle, color=cyan!60!black},    
        g={mark=triangle, color=black},        
        h={mark=triangle, color=teal},     
        i={mark=triangle, color=magenta},         
        j={mark=triangle*, color=brown}           
    },  xlabel={CIFAR-10}, ylabel={TinyImageNet}, label style={font=\scriptsize},
        after end axis/.code={
    \node[anchor=north, font=\small] at (current axis.north) {(b)};
},
]

\addplot[scatter,only marks,scatter src=explicit symbolic]
table[meta=label] {
    x    y     label
    81.70 81.10  a
    67.60 58.00  b
    89.10 69.30  d
    89.50 74.80  e
    97.30 90.70  f
    91.00 78.20  g
    80.30 77.20  h
    91.30 82.30  i
};

\node at (axis cs: 81.70, 81.10) [anchor=south] {\tiny OpenMax \cite{bendale2016towards}};
\node at (axis cs: 67.60, 58.00) [anchor=west] {\tiny G-OpenMax \cite{ge2017generative}};
\node at (axis cs: 89.10, 69.30) [anchor=west] {\tiny PROSPER \cite{zhou2021learning}};
\node at (axis cs: 89.50, 74.80) [anchor=north] {\tiny C2AE \cite{oza2019c2ae}};
\node at (axis cs: 97.30, 90.70) [anchor=east] {\tiny OpenGAN \cite{kong2021opengan}};
\node at (axis cs: 91.00, 78.20) [anchor=east] {\tiny ARPL \cite{chen2021adversarial}};
\node at (axis cs: 80.30, 77.20) [anchor=north] {\tiny CAC \cite{miller2021class}};
\node at (axis cs: 91.30, 82.30) [anchor=south] {\tiny CSSR \cite{huangClassSpecificSemanticReconstruction2022}};

\nextgroupplot[
        xmajorgrids = false,
    ymajorgrids = false,
    scatter/classes={
        b={mark=triangle, color=red},    
        d={mark=triangle, color=orange},
        e={mark=triangle, color=violet},    
        f={mark=triangle, color=cyan!60!black},    
        g={mark=triangle, color=black},        
        h={mark=triangle, color=teal},     
        i={mark=triangle, color=magenta},         
        j={mark=triangle*, color=brown}           
    },  xlabel={Near-OOD AUROC}, ylabel={Far-OOD AUROC}, label style={font=\scriptsize},
    after end axis/.code={
    \node[anchor=north, font=\small] at (current axis.north) {(c)};
},
]

\addplot[scatter,only marks,scatter src=explicit symbolic]
table[meta=label] {
    x    y     label
    93.10 95.94  b
    92.33 96.74  d
    94.82 96.00  e
    91.03 91.00  f
    94.86 98.18  g
    89.51 92.00  h
    88.73 91.93  i
    89.43 91.66  j
};

\node at (axis cs: 93.10, 95.94) [anchor=west] {\tiny PixMix \cite{hendrycks2022pixmix}};
\node at (axis cs: 92.33, 96.74) [anchor=west] {\tiny Logitnorm \cite{wei2022mitigating}};
\node at (axis cs: 94.82, 96.00) [anchor=north] {\tiny OE \cite{hendrycks2018deep}};
\node at (axis cs: 91.03, 91.00) [anchor=west] {\tiny MCD \cite{zou2024mcd}};
\node at (axis cs: 94.86, 98.18) [anchor=east] {\tiny RotPred \cite{hendrycks2019using}};
\node at (axis cs: 89.51, 92.00) [anchor=south] {\tiny CSI \cite{tack2020csi}};
\node at (axis cs: 88.73, 91.93) [anchor=north] {\tiny MixOE \cite{zhang2023mixture}};
\node at (axis cs: 89.43, 91.66) [anchor=west] {\tiny AugMix \cite{hendrycks2019augmix}};


\nextgroupplot[
        xmajorgrids = false,
    ymajorgrids = false,
    scatter/classes={
        a={mark=triangle,   color=blue},       
        c={mark=triangle, color=green!60!black}, 
        d={mark=triangle, color=orange},
        e={mark=triangle, color=violet},    
        f={mark=triangle, color=cyan!60!black},    
        g={mark=triangle, color=black},        
        h={mark=triangle, color=teal},     
        i={mark=triangle, color=magenta},         
        j={mark=triangle*, color=brown}           
    },     xlabel={Class-IL},
    ylabel={Task-IL},label style={font=\scriptsize},
    after end axis/.code={
    \node[anchor=north, font=\small] at (current axis.north) {(d)};
},
]

\addplot[scatter,only marks,scatter src=explicit symbolic]
table[meta=label] {
    x    y     label
    44.79 91.19  a
    23.80 67.82  c
    30.74 75.92  d
    66.35 92.66  e
    24.50 90.70  f
    27.92 92.75  g
    94.60 99.43  h
    61.93 91.40  i
    65.57 93.43  j
};

\node at (axis cs: 44.79, 91.19) [anchor= south west] {\tiny ER \cite{rolnick2019experience}};
\node at (axis cs: 23.80, 67.82) [anchor=south] {\tiny iCaRL \cite{rebuffi2017icarl}};
\node at (axis cs: 30.74, 75.92) [anchor=east] {\tiny CVT \cite{wang2022online}};
\node at (axis cs: 66.35, 92.66) [anchor=west] {\tiny SCoMMER \cite{sarfraz2023sparse}};
\node at (axis cs: 24.50, 90.70) [anchor=north west] {\tiny DualNet \cite{pham2021dualnet}};
\node at (axis cs: 27.92, 92.75) [anchor=south] {\tiny BiMeCo \cite{nie2023bilateral}};
\node at (axis cs: 94.60, 99.43) [anchor=north] {\tiny ICL \cite{man2024icl}};
\node at (axis cs: 61.93, 91.40) [anchor=north west] {\tiny DER \cite{buzzega2020dark}};
\node at (axis cs: 65.57, 93.43) [anchor=south] {\tiny Co2L \cite{cha2021co2l}};


\end{groupplot}
\end{tikzpicture}
\caption{{Results overview of existing popular benchmarks in mitigating Concept Shift. (a) Plot of AUROC vs AUCPR values of existing AD approaches. (b) AUROC plot of OSR methods against two popular benchmark datasets: CIFAR-10 and TinyImageNet. (c) Near and Far OOD AUROC plot of popular OOD detection benchmarks. (d) Accuracy plot of Task-Incremental Learning vs Class-Incremental Learning for popular CL benchmarks.}}
\label{fig:semantic_benchmark}
\end{figure*}
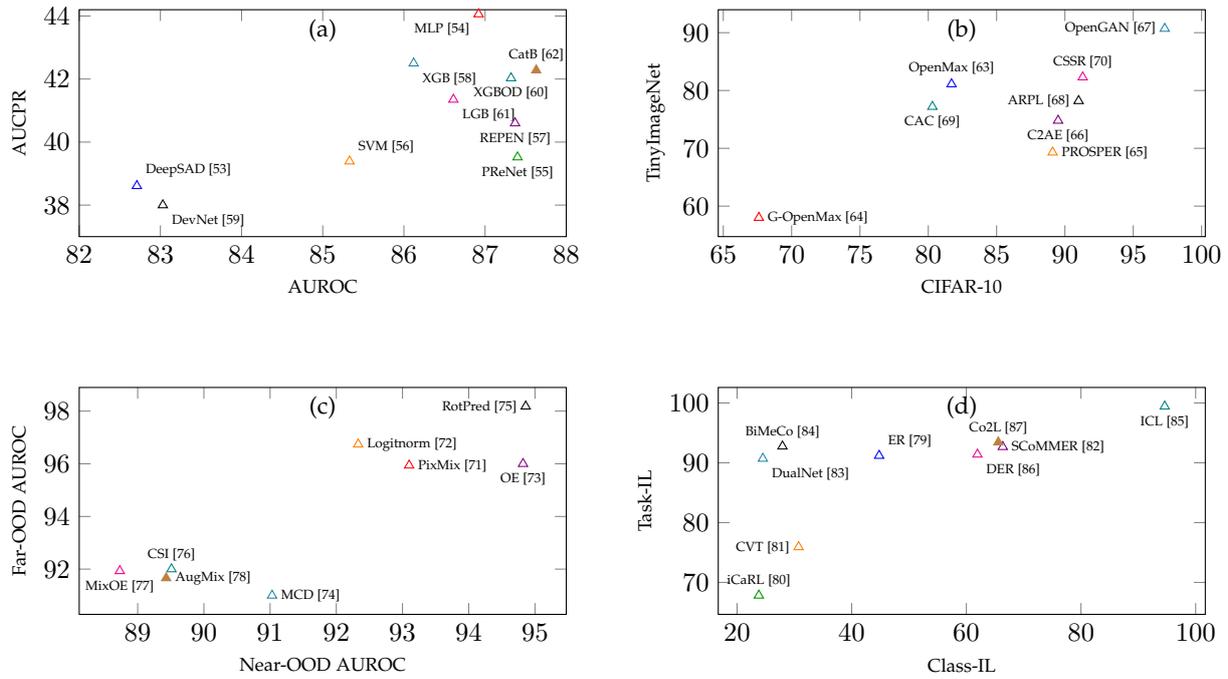

\noindent Our study offers a comprehensive survey of existing methodologies for mitigating data distribution shifts. The paper is structured to guide readers systematically, ensuring optimal comprehension. We discuss, compare, and report benchmark results for each mitigation strategy presented in the taxonomy illustrated in Fig. \ref{fig:taxonomy_one}. Initially, we provide a systematic comparison of methodologies, as depicted in Tables \ref{table:covariate_comp} and \ref{table:semantic_comp}. This comparison is conducted from a fundamental perspective, evaluating different criteria pertinent to each shift type. Subsequently, we present schematic diagrams of each mitigation methodology in Figs. \ref{fig:covariate_schematic} and \ref{fig:concept_schematic}, illustrating the operational mechanisms of these methods concerning data-point/sample classification and decision boundary establishment. Thirdly, we report benchmark results of methodologies across covariate and semantic shifts in Figs. \ref{fig:covariate_benchmark} and \ref{fig:semantic_benchmark}. {Here, we present a comparative overview of the highest reported performance metrics (e.g., accuracy, AUROC) for each method on the target dataset, as documented in their respective papers. All values are drawn directly from the original sources, without re-evaluation. Furthermore, to enhance practical guidance for the readers, we also discuss most of these methods in a separate tables (Tables. \ref{tab:cov_proscons}, and  \ref{tab:sem_proscons}) where we point out their core working strategy, best use case, and potential limitations. Lastly, through Tables \ref{tab:covariate_applied}, and \ref{tab:concept_applied}, we provide a comprehensive report of latest research in several applied domains, highlighting corresponding shift handling mechanisms, along with their core technological synopsis.}
\subsection{Covariate/Feature Shifts}
\noindent The phenomenon where the distribution of input features (or covariates) in the training data diverges from that in the test data test while the conditional distribution of the targets given the inputs remains unchanged is known as distribution shift in the feature space \cite{sugiyama2007covariate}. This shift can be particularly troublesome in real-world situations where the context or environment in which models are used changes over time or is not the same as the one in which they were trained. Neglecting these changes may result in less than ideal model performance or even model failure.
\par As investigated by \cite{arjovsky2019invariant}, the models trained on ERM often use a cheating way to perform classification, by learning spurious features from the training data which holds no stable properties of the sample. Under covariate shift, often models are very likely to pick up these spurious correlation while missing out the robust features that has causal relationship with the output labels \cite{zhou2021examining}. This inherent fact often gives rise to poor generalization on the new data that are sampled without such biases.

\begin{figure}[t]
\centering
\begin{tikzpicture}
\begin{groupplot}[
    group style={
        group size=1 by 3,
        horizontal sep=2cm,
    },
    width=0.45\textwidth,
    height=0.25\textwidth,
    grid=both
]
\nextgroupplot[
    ylabel={Accuracy},
    ymin=84, ymax=98,
    symbolic x coords={SAM,$\mu$Net,BigT,Bamboo,Astro,CeiT,GPIPE,ASANas},
    xtick=data,
    x tick label style={font=\scriptsize, rotate=45, anchor=east},
    enlarge x limits=0.1,
    ybar,
    bar width=6pt,
    after end axis/.code={
        \node[anchor=north, font=\small] at (current axis.north) {(a)};
    },
    nodes near coords={},
    every node near coord/.append style={text=}
]
\addplot+[] coordinates {
    (SAM, 96.08)
    ($\mu$Net, 94.95)
    (BigT, 93.5)
    (Bamboo, 90.2)
    (Astro, 93.36)
    (CeiT, 91.8)
    (GPIPE, 91.3)
    (ASANas, 85.42)
};

\node at (axis cs:SAM,96.08) [font=\tiny, anchor=south] {\cite{foret2020sharpness}};
\node at (axis cs:$\mu$Net,94.95) [font=\tiny, anchor=south] {\cite{gesmundo2022evolutionary}};
\node at (axis cs:BigT,93.5) [font=\tiny, anchor=south] {\cite{kolesnikov2020big}};
\node at (axis cs:Bamboo,90.2) [font=\tiny, anchor=south] {\cite{zhang2022bamboo}};
\node at (axis cs:Astro,93.36) [font=\tiny, anchor=south] {\cite{dagli2023astroformer}};
\node at (axis cs:CeiT,91.8) [font=\tiny, anchor=south] {\cite{yuan2021incorporating}};
\node at (axis cs:GPIPE,91.3) [font=\tiny, anchor=south] {\cite{huang2019gpipe}};
\node at (axis cs:ASANas,85.42) [font=\tiny, anchor=south] {\cite{macko2019improving}};

\nextgroupplot[
    ylabel={Accuracy},
    ymin=65, ymax=95,
    symbolic x coords={CDAN,SWG,PGA,MDD,GVB,FixBi,SHOT,FDA,GSDE,ELS},
    xtick=data,
    x tick label style={font=\scriptsize, rotate=45, anchor=east},
    enlarge x limits=0.1,
    ybar,
    bar width=6pt,
    after end axis/.code={
        \node[anchor=north, font=\small] at (current axis.north) {(b)};
    },
    nodes near coords={},
    every node near coord/.append style={text=}
]
\addplot+[] coordinates {
    (CDAN, 70.0)
    (SWG, 84.3)
    (PGA, 75.8)
    (MDD, 72.2)
    (GVB, 75.3)
    (FixBi, 78.7)
    (SHOT, 81.8)
    (FDA, 90.1)
    (GSDE, 73.6)
    (ELS, 84.6)
};

\node at (axis cs:CDAN,70.0) [font=\tiny, anchor=south] {\cite{long2018conditional}};
\node at (axis cs:SWG,84.3) [font=\tiny, anchor=south] {\cite{westfechtel2023combining}};
\node at (axis cs:PGA,75.8) [font=\tiny, anchor=south] {\cite{phan2024enhancing}};
\node at (axis cs:MDD,72.2) [font=\tiny, anchor=south] {\cite{li2020maximum}};
\node at (axis cs:GVB,75.3) [font=\tiny, anchor=south] {\cite{cui2020gradually}};
\node at (axis cs:FixBi,78.7) [font=\tiny, anchor=south] {\cite{na2021fixbi}};
\node at (axis cs:SHOT,81.8) [font=\tiny, anchor=south] {\cite{liang2020we}};
\node at (axis cs:FDA,90.1) [font=\tiny, anchor=south] {\cite{yang2020fda}};
\node at (axis cs:GSDE,73.6) [font=\tiny, anchor=south] {\cite{westfechtel2024gradual}};
\node at (axis cs:ELS,84.6) [font=\tiny, anchor=south] {\cite{saporta2020esl}};

\nextgroupplot[
    scatter/classes={
        a={mark=triangle,   color=blue},       
        b={mark=triangle, color=red},    
        c={mark=triangle, color=green!60!black}, 
        d={mark=triangle, color=orange},
        e={mark=triangle, color=violet},    
        f={mark=triangle, color=cyan!60!black},    
        g={mark=triangle, color=black},        
        h={mark=triangle, color=teal},     
        i={mark=triangle, color=magenta},         
        j={mark=triangle*, color=brown}           
    },    xlabel={VLCS},
    ylabel={PACS}, label style={font=\scriptsize},
    after end axis/.code={
    \node[anchor=north, font=\small] at (current axis.north) {(c)};
},
]

\addplot[scatter,only marks,scatter src=explicit symbolic]
table[meta=label] {
    x    y     label
    77.5 85.5  a
    78.5 83.5  b
    78.6 83.6  c
    77.5 84.6  d
    77.4 84.6  e
    78.9 87.1  f
    82.4 98.6  g
    79.9 88.6  h
    82.4 97.0  i
    79.2 85.6  j
};

\node at (axis cs: 77.5, 85.5) [anchor= south west] {\tiny ERM \cite{gulrajani2020search}};
\node at (axis cs: 78.5, 83.5) [anchor=south west] {\tiny IRM \cite{arjovsky2019invariant}};
\node at (axis cs: 78.6, 83.6) [anchor= west] {\tiny DANN \cite{ganin2015unsupervised}};
\node at (axis cs: 77.5, 84.6) [anchor=west] {\tiny MMD \cite{li2018domain}};
\node at (axis cs: 77.4, 84.6) [anchor=north] {\tiny MixUp \cite{zhang2017mixup}};
\node at (axis cs: 78.9, 87.1) [anchor=south] {\tiny SWAD \cite{cha2021swad}};
\node at (axis cs: 82.4, 98.6) [anchor=east] {\tiny PromptStyler \cite{cho2023promptstyler}};
\node at (axis cs: 79.9, 88.6) [anchor=south] {\tiny SIMPLE \cite{li2022simple}};
\node at (axis cs: 82.4, 97.0) [anchor=east] {\tiny SoftPrompt \cite{bai2024soft}};
\node at (axis cs: 79.2, 85.6) [anchor=west] {\tiny GMDG \cite{tan2024rethinking}};

\end{groupplot}
\end{tikzpicture}
\caption{{Results overview of existing popular benchmarks in mitigating Covariate Shift. (a) Transfer Learning accuracies on CIFAR-100 datasets, (b) Domain Adaptation accuracies on OfficeHome dataset, and (c) Domain Generalization accuracies on PACS and VLCS datasets.} }
\label{fig:covariate_benchmark}
\end{figure}
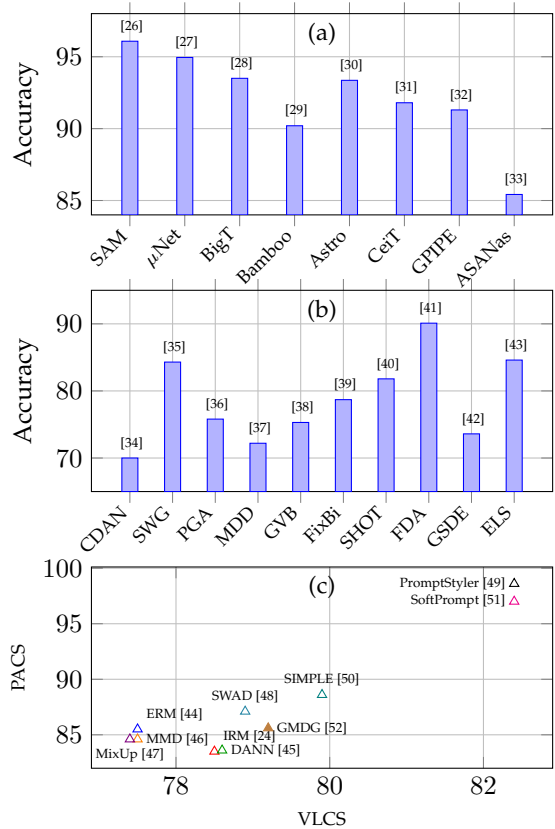

\subsubsection{Transfer Learning}
\noindent Transfer learnings are often used from the data sufficient source task to complement the similar but non-identical target task with limited training samples \cite{zhuang2020comprehensive}. The objective of transfer learning technique is to reduce the amount of new labeled data required in the target domain, and possibly avoid the cost of collecting an entire new labeled training data\cite{wang2014active}. In the context of transfer learning, a domain is defined by its feature space and its marginal probability distribution while a task is characterized by its label space and an associated objective predictive function. \textit{Transductive transfer learning} \cite{joachims1999transductive} explains the phenomenon similar to that of Domain Adaptation, where there is shift in the domains of training and test sets without the task being changed. In this kind of setting, it is possible that either the feature spaces are different or the marginal probability distributions of the input data are different \cite{pan2009survey}. \par In the past, transfer learning approaches considered specific parts of the model to be carried over between tasks \cite{do2005transfer}, \cite{raina2006constructing}, until recently where large cohort of researches \cite{zamir2018taskonomy}, \cite{long2017deep} focused on the problem of data distribution changes, especially relating to the covariate shift. A study in \cite{neyshabur2020being} actually carried out an investigation to answer what knowledge is being transferred from the source domain to the target domain in the process of transfer learning. This study offered novel tools and analysis approach to identify factors that contribute to successful transfer and pinpoint the network components responsible for it.  \cite{bengio2012deep} delved into the context of the Unsupervised Transfer Learning Challenge, highlighting the benefits of unsupervised pre-training of representations and demonstrating how it can be leveraged in situations where the focus is on generalizing to the instances that originate from a different distribution than the training set. Other variants have extended the fundamental idea of this field into adaptive \cite{cao2010adaptive} transfer learning that deals with learning an accurate model using tiny amount of new data, and online transfer learning \cite{zhao2014online} which makes an assumption of training data in the new domain arriving sequentially. 
\par A TL algorithm to handle both support and model shift was studied in \cite{wang2014flexible}. In this algorithm, the shift handling is performed by transforming both features, and labels of the input by a location-scale shift allowing more flexible transformations. In \cite{liu2021deep}, a conditional shift regression task was studied using deep transfer learning for machine health monitoring in industrial application. The authors specifically proposed a hybrid loss function with achieve two objectives; reducing the prediction error, and preserving global characteristics of conditional distribution dominated by target data. A similar study in TL for regression under conditional shift was conducted in \cite{chen2020transfer} where they considered a special case of source and target domains sharing same margin distributions but non-identical conditional probability distributions. The authors proposed a framework for TL based on fuzzy residual that can learn the target model in a model agnostic way without neglecting the properties of the source data. 
\par The literature in the use case of transfer learning has been substantial, and is not plausible to cover all of them in detail. 
Therefore, we guide our readers to other applied studies that relays specific use case of transfer learning in reinforcement learning \cite{zhu2023transfer}, medical image analysis \cite{kim2022transfer}, machinery fault diagnosis \cite{li2022perspective}, sentiment analysis \cite{chan2023state}, intrusion detection systems \cite{mehedi2022dependable}. 
\subsubsection{Domain Adaptation}
\noindent The statistical attributes of data from any domain might undergo transformations over time, or newly acquired samples could accumulate from diverse sources, leading to what is known as domain shift. When there is a misalignment between the distributions of training and test data, the performance of the trained model is prone to deteriorate upon application to the test data. Domain adaptation (DA) represents a specific subset of transfer learning where labeled data from one or multiple pertinent source domains is leveraged to perform tasks in a distinct target domain \cite{wang2018deep}. The principal objective of domain adaptation is to learn a model using labeled data from the source domain that can generalize well to the target domain by minimizing the disparities between the domain distributions\cite{farahani2021brief}. There has been numerous study in the long line of literature in the field of supervised DA \cite{motiian2017unified}, \cite{saha2011active}, \cite{abdelwahab2015supervised}, \cite{koniusz2018museum}, semi-supervised DA \cite{saito2019semi}, \cite{daume2010frustratingly}, \cite{he2020classification} and unsupervised DA \cite{ganin2015unsupervised}, \cite{sun2016return}, \cite{kang2019contrastive}, \cite{li2020model} attempting to solve the non-trivial task of adapting the source trained model into the target domain in ML systems. Specifically, in this section we try to explain the DA techniques specifically curated for dealing with covariate distribution shift. {While our taxonomy groups methods according to their primary design focus, e.g., DA methods under covariate shift it is important to acknowledge that many modern DA variants inherently address aspects of semantic shift as well. Variants such as open-set DA \cite{panareda2017open}, partial DA \cite{cao2018partial}, universal DA \cite{you2019universal}, and class-incremental DA \cite{wulfmeier2018incremental} explicitly account for label space mismatches, including scenarios where the target domain contains unseen or partially overlapping classes. These approaches extend beyond the classical covariate shift assumption (i.e., $P(X)$ changes while $P(Y|X)$ remains fixed) and instead operate in regimes where the conditional distribution $P(Y|X)$ itself shifts, which is central to semantic shift. However, these methods typically address semantic shift arising from label space divergence, rather than deeper semantic reinterpretations of same label.}
\par In \cite{zhang2013domain}, researchers have considered a DA study in handling two types of distribution shift; one being the distribution of the covariates, and other the conditional distribution of the target data given cross domain covariate shift. To handle such shifts, the study proposed approaches based on kernel mean embedding of distributions (conditional and marginal), empirically verifying their theoretical claims with experiments on real world problems. Another research studied the DA problem under open set label shift where label distribution can change unexpectedly as well as novel concepts can appear during deployment \cite{garg2022domain}. They proposed learning a target classifier, termed as Positive-Unlabeled (PU) learning where the learners objective is to estimate the target label distribution including that of newly introduced classes as well. With rigorous experiments across several vision, medical, and language benchmark datasets, a well-posed problem was offered with significant improvement in the target domain accuracy. Similarly, a large-scale benchmark was introduced in \cite{garg2023rlsbench} that consisted of over 500 distribution shift pairs across language, vision, tabular datasets. These distribution shifts focused not only on class-conditional shifts but also the label marginal shifts. 
\subsubsection{Domain Generalization}
\noindent In practical scenarios, it's infeasible to collect training data across every conceivable domain. The capacity of a model to extrapolate from familiar domains to unfamiliar ones is paramount. Domain generalization addresses the intricate task of educating a model using data from one or several source domains so that it can adeptly generalize to novel, unseen target domains that share the same label space. Very often in the literature \cite{qiao2020learning}, \cite{gulrajani2020search}, \cite{ye2021towards} this technique is used to in conjunction with out of distribution (OOD) generalization \cite{krueger2021out}. The only difference being the former employs multiple training datasets from different domain for model training purpose \cite{spratling2023comprehensive}. Nevertheless, this is achieved without the luxury of accessing any data from the target domain during the training phase. While the OOD generalization is a more generic term, both share the same objective of generalizing well on unseen domain by capturing domain-agnostic representations.
\par In the realm of machine learning, the generalization aptitude of a model is frequently contingent upon the volume and heterogeneity of the training dataset. When confronted with a constrained dataset, data augmentation emerges as one of the most cost-effective and straightforward strategies to proliferate samples, thereby bolstering the model's generalizability. The primary aim of data augmentation-driven techniques is to amplify the variance within the existing training dataset by employing diverse data manipulation methodologies. Concurrently, this process also augments the overall volume of the dataset.
One of the DG study in \cite{chuang2020estimating} aimed to improve the performance estimation of the model in the presence of distributional shift without supervision. They used a set of domain-invariant representations as a proxy model for an unknown true target labels where the accuracy of the resulting risk estimates depended on the target risk of that model. The study addressed the generalization of range-invariant representations and showed that the complexity of the latent representation has a significant impact on target risk. Empirically, their method facilitated self-tuning of the DA models while accurately estimating the target error of a given model under distributional shifts. Another empirical paper \cite{ding2021closer} studied the problem of graph OOD generalization by evaluating eight different datasets representing different types of distributional shifts on graphs. These datasets were used to perform a comprehensive empirical evaluation of popular DG algorithms, graph expansion methods and GNN models. They came up with an interesting deduction that most DG algorithms did not improve OOD generalization performance when confronted with different types of domain shifts on the graph. Instead they discovered that the optimal combination of advanced GNN models and robust graph expansion methods can effectively achieve state-of-the-art performance in graph OOD generalisation problem.
\par Unlike the straightforward application of principle of invariance in images, identifying invariant features within graph data is inherently challenging. A study in \cite{chen2022learning} addressed this challenge of applying invariance principles to graph data under distribution shifts. In particular, they proposed a causality inspired invariant graph learning to ensure OOD generalization on graphs. The authors claim that true OOD generalization can be achievable if the focus is shifted towards subgraphs that hold substantial information regarding the causality behind label assignments. The proposed technique involved an information-theoretic objective designed to identify and safeguard these invariant intra-class information, thereby ensuring that the learned subgraph representations are resilient to distribution shifts. 

\begin{table*}[htbp]
\centering
\small
\setlength{\tabcolsep}{5pt}
\renewcommand{\arraystretch}{1.35}
\rowcolors{2}{gray!5}{white}
\caption{List of Popular Covariate Shift Mitigation Papers, including their Core Strategy, Best use case, and Limitations.}
\label{tab:cov_proscons}

\resizebox{\textwidth}{!}{%
\begin{tabular}{>{\bfseries}l l p{5.5cm} p{5.5cm} p{5.5cm}}
\rowcolor{gray!20}
\textbf{Type} & \textbf{Reference (Year)} & \textbf{Core Strategy} & \textbf{Best use} & \textbf{Limitation} \\
\midrule

& SAM \cite{foret2020sharpness} (2020) & Minimizes sharpness and loss jointly & Reduces overfitting to source domains & High compute cost; ignores domain structure \\
& $\mu$Net. \cite{gesmundo2022evolutionary} (2022) & Evolutionary search over modular subnets & Modular transfer across domains & Costly search; limited to predefined modules\\
TL
& BigT \cite{kolesnikov2020big} (2020) & Transfer of large pre-trained models & Strong baseline under covariate shift & Requires massive compute and data resources \\
& Bamboo \cite{zhang2022bamboo} (2022) & Online model selection with bootstrapped risk & Dynamic adaptation under shift & Needs access to unlabeled target data streams \\
& Astro \cite{dagli2023astroformer} (2023) & Transformer-based meta-learned initialization & Few-shot adaptation under distribution shift & High model complexity; meta-training required \\
& CeiT \cite{yuan2021incorporating} (2021) & Combines CNN token embedding with Vision Transformer & Visual tasks needing inductive bias & Less effective on small data or non-visual domains \\

\midrule
& CDAN \cite{long2018conditional} (2018) & Aligns joint feature–label distributions via adversarial training & Supervised DA with label shift & Sensitive to classifier confidence and adversarial training stability \\
& SWG \cite{westfechtel2023combining} (2023) & Aligns Wasserstein geometry between domains & Unsupervised DA with strong structure shift & Computationally intensive; limited scalability\\
& PGA \cite{phan2024enhancing} (2024) & Promotes geometry alignment using pseudo labels and self-training & Unsupervised DA with label imbalance & Relies on pseudo label quality; sensitive to noise \\
DA
& MMD \cite{li2020maximum} (2020) & Minimizes domain discrepancy via MMD loss & Simple and effective DA baseline & Simple and effective DA baseline \\
& GVB \cite{cui2020gradually} (2020) & Gradually aligns features via auxiliary boundary loss & DA with class-boundary refinement & Requires careful scheduling and tuning \\
& FixBi \cite{na2021fixbi} (2021) & Bidirectional self-training with reliable pseudo labels & Class-balanced unsupervised DA & Sensitive to early pseudo label errors \\
& SHOT \cite{liang2020we} (2020) & Source-free DA using feature clustering and pseudo labeling & DA when source data is unavailable & Relies on target structure; tuning is tricky \\

\midrule
& ERM \cite{gulrajani2020search} (2020) & Minimizes average empirical risk across source domains & Strong baseline for DG & Ignores domain-specific signals and variance \\
& IRM \cite{arjovsky2019invariant} (2019) & Learns invariant predictors across domains & DG with strong causal assumptions & Hard to optimize; often underperforms in practice \\
& DANN \cite{ganin2015unsupervised} (2015) & Adversarial feature alignment via domain classifier & Early DA and DG benchmark & May confuse domain-invariant and task-relevant features \\
DG
& MixUp \cite{zhang2017mixup} (2017) & Interpolates inputs and labels for regularization & Improves generalization and robustness & May underperform on complex or structured shifts \\
& SWAD \cite{cha2021swad} (2021) & Averages weights from flat minima for stable generalization & DG under training instability & Assumes flatness correlates with generalization \\
\bottomrule
\end{tabular}%
}
\end{table*}

\begin{table*}[htbp]
\centering
\small
\setlength{\tabcolsep}{5pt}
\renewcommand{\arraystretch}{1.35}
\rowcolors{2}{gray!5}{white}
\caption{List of Applied Research Papers Dealing with Covariate Shift Problem}
\label{tab:covariate_applied}

\resizebox{\textwidth}{!}{%
\begin{tabular}{>{\bfseries}l l l p{5.5cm} l l}
\rowcolor{gray!20}
\textbf{Type} & \textbf{Reference (Year)} & \textbf{Applied Area} & \textbf{Core Technology} & \textbf{Dataset Type} & \textbf{Used Metrics} \\
\midrule

& Sohn et al. \cite{sohn2023visual} (2023) & Image Synthesis & Generative vision transformers, Prompt tuning & Vision data & FID, LPIPS \\
& Qian et al. \cite{qian2023deep} (2023) & Machine Fault Diagnosis & Conditional alignment, I-Softmax loss & Vibration data & Accuracy \\
TL
& Zhu et al. \cite{zhu2023bayesian} (2023) & Machine Fault Diagnosis & Bayesian semi-supervised TL, MC dropout & Vibration data & RMSE, MAE \\
& Bierbrauer et al. \cite{bierbrauer2023transfer} (2023) & Intrusion Detection & 1D CNN, Random Forest & Network traffic & TPR, FPR, TNR \\
& Zhou et al. \cite{zhou2024spatial} (2024) & Intelligent Transport & Federated TL, Siamese NN, Spatio-temporal clustering & Positioning data & RMSE, MAE, MAPE \\
& Xiao et al. \cite{xiao2024selective} (2024) & Heterogeneous Labels & Random Walk, LSTM, Meta-learning & Image data & Transfer Accuracy \\

\midrule
& Li et al. \cite{li2024subspace} (2024) & Image Classification & Variational NN, Conditional alignment & Synthetic, Multi-domain & Adaptation Accuracy \\
& Hoyer et al. \cite{hoyer2023mic} (2023) & Visual Recognition & Unsupervised DA, Masked image modeling & Segmented images & Accuracy, IoU \\
& Truong et al. \cite{truong2023fredom} (2023) & Scene Understanding & Conditional structure net, Self-attention & Vision data & Accuracy \\
DA
& Kim et al. \cite{kim2023datid} (2023) & Text-to-Image & CLIP, Diffusion, Pose filtering & Text-image, 3D data & KID \\
& Wang et al. \cite{wang2023class} (2023) & Text Understanding & Conditional alignment, Optimal transport & Multi-domain image data & Accuracy \\
& Chen et al. \cite{cheng2023adpl} (2023) & Semantic Segmentation & Dual-path translation, ClassMix & Landscape images & Accuracy, IoU \\
& Hao et al. \cite{hao2023dual} (2023) & Video-Text Retrieval & Dual alignment, Cross-modal embedding & Video-text pairs & Median ranking \\
& Ge et al. \cite{ge2023domain} (2023) & Image Classification & CLIP, Prompt learning, Domain embedding & Text-image pair & Accuracy \\

\midrule
& Wang et al. \cite{wang2023sharpness} (2023) & Image Classification & Sharpness-aware gradient matching & Image data & Accuracy (ID/OOD) \\
& Segu et al. \cite{segu2023batch} (2023) & Image Classification & Batch norm variants, Latent space learning & Image data & Generalization Accuracy \\
& Chen et al. \cite{chen2023federated} (2023) & Image Recognition & Federated learning, Adaptive normalization & Image data & Generalization Accuracy \\
DG
& Yu et al. \cite{yu2023mind} (2023) & Node Classification & Label-invariant augmentation, GNN & Synthetic graph data & Generalization Accuracy \\
& Zhang et al. \cite{zhang2023map} (2023) & Image Classification & Bilevel optimization, Adapter layers & Image data & Generalization Accuracy \\
& Wang et al. \cite{wang2023out} (2023) & Data Mining & Conditional independence test, Causal selection & Synthetic, Medical image & RMSE, Avg. error \\

\bottomrule
\end{tabular}%
}
\end{table*}

\subsection{Concept/ Semantic Shifts}
\noindent \cite{mitchellMachineLearning2013} defines concept as a function learned by an algorithm that maps input values to their corresponding output values, as defined by a set of training examples. Concept shift, or drift in some literature \cite{tsymbal2004problem} occurs when the posterior probabilities of the input and labels change. Due to the complex and dynamic nature of data distribution shift that occur over time, the model can be presented with new concepts (e.g. new categories of objects) at any time \cite{hsu2020generalized}. The introduction of new concepts can result in the catastrophic failure of a model due to its reliance on the iid hypothesis for prediction. Therefore, handling these shifts is essential for maintaining the robustness of ML models. Often in computer vision, the change of concept is used interchangeably with change in semantics as it represents the change in features intrinsic to the object \cite{tian2021exploring}, \cite{yang2021generalized}. 
\par In this section, we outline various methodologies and techniques that have been studied as part of addressing this problem, with either explicit or implicit association with the concept shift.

\subsubsection{Open Set Recognition}
\noindent Traditional classification methods require the system to classify all the test instances into one of the trained classes, disregarding the prevalence of concept shift that takes places in an open world. Instead a robust ML system must constrain its classification criteria within the known paradigm of learned classes whilst rejecting unseen classes that are irrelevant and meaningless to what it has learned. Open set recognition holds two supposition to enhance the robustness of ML systems; one is to accurately classify samples into known categories, and the other is to detect and reject unknown samples \cite{geng2020recent}. Sometimes also referred as open world machine learning \cite{parmar2023open}, this approach aims to eliminate the risk of mistakenly categorizing an unknown instance into one of the known categories. One of the challenges in formulating OSR method is to optimize the model for accurately estimating the probability of all known classes while maintaining precise recognition of the unknown classes \cite{mahdaviSurveyOpenSet2021}. In this regard, several approaches have been studied in ML research to address this concept shift problem.
\par In general, majority of DNN's penultimate layer are connected to the Softmax layer which is responsible to produce a probability distribution over the total number of known concepts it is trained on \cite{krizhevsky2012imagenet}, \cite{chatfield2014return}. It was a common technique for handling samples arising from unknown concept by assigning a threshold with an assumption that unknown samples would incur low probability. However, this type of uncertainty thresholding technique was later found to be simply not enough to determine what is unknown because of two reasons. First, the unknown samples are usually known to hold a really large sample space in an open world and generalizing to this large subspace was challenging. {Second, the counterfactual, and adversarial images can really fool the model by producing high confidence scores regardless of them being unknown}. To cope up with this issues, one of the preliminary study in OSR proposed OpenMax \cite{bendale2016towards} where the existing deep neural network (DNN) was modified by introducing a new layer with the objective of assessing the likelihood of the input belong to an unknown concept. By applying distance normalization process based on extreme-value meta recognition on the activation patterns of DNN's penultimate layer, the rejection probability was determined for unknown images. By doing so the system was able to effectively reject misleading, unknown, and even many adversarial images significantly reducing the obvious error of traditional DNN's in open space.
Several consecutive studies were studied aiming to enhance the OSR benchmark results. An interesting OSR research based on sparse representation was studied in \cite{zhang2016sparse} which used class reconstruction errors for classification task. The proposed framework is grounded on the principle of Extreme Value Theory (EVT) and unfolds in two main stages. The approach initially models the tail distributions of both matched and non-matched reconstruction errors by employing EVT, thereby transforming the intricate OSR issue into two separate hypothesis testing situations. Afterwards, in the second phase, the method entails calculating the reconstruction errors for a test sample from each category and the confidence scores that originate from the two tail distributions are combined to identify the test sample's actual identity. 
\subsubsection{Out-of-Distribution Detection}
\noindent The emergence of out-of-distribution (OOD) detection \cite{tamang2024margin} in the field of deep learning is a response to the common issue of models being overconfident in classifying samples from different semantic distributions in image classification and text categorization tasks \cite{yang2021generalized}. This methodology relies on a scoring function that converts the input into an OOD score, signifying the extent to which the sample is considered differently distributed from that of training data \cite{zhang2023decoupling}. Although, the separation of ID and OOD data remains a non trivial task, it is arguable that continued research progress in OOD detection requires insights into the fundamental cause and mitigation of model overconfidence on OOD \cite{sun2021react}.

\par In \cite{huang2021importance}, the researchers introduced GradNorm, a simple yet effective method that utilizes information from the gradient space to identify OOD inputs. GradNorm specifically utilized vector norm of gradients, which are backpropagated from the Kullback-Leibler divergence between the softmax output and a uniform probability distribution. The key assumption underlying this approach is that the magnitude of gradients is generally greater for ID data than for OOD data. This characteristic makes the gradient magnitude a useful metric for detecting OOD inputs. In another study in \cite{wu2023meta} the researchers put forward continuously adaptive out-of-distribution (CAOOD) detection framework that was developed with the intention of creating a model that could rapidly adapt to new distributions, especially when there are insufficient ID samples available during deployment. Specifically, the authors devised a meta out-of-distribution learning (MOL) strategy which involved creating a 'learning-to-adapt' diagram that facilitates the initial learning of an effectively initialized OOD detection model during the training phase. During the testing phase, MOL aimed to maintain the OOD detection's efficiency across varying distributions by allowing for swift adaptation to new distributions through minimal adjustments.

\par {One of the core difficulties in OOD detection is that OOD inputs can be extremely diverse, and without any assumptions, detecting anything that’s not ID is provably impossible. In other words, if we place no restrictions on what the OOD data could be, no finite training procedure can guarantee detection of every possible OOD input. Intuitively, an algorithm that works well for one type of unseen data can always be fooled by another type, unless we have some prior knowledge or constraints. Therefore, theoretical analyses of OOD detection \cite{morteza2022provable, fang2022out, ovadia2019can} introduce explicit assumptions or models of the data to make the problem tractable. A common assumption is that the ID and OOD distributions are sufficiently distinct in some feature space (for example, they may have disjoint support or minimal overlap). If OOD examples can occupy the same feature regions as ID ones, then no detector can perfectly separate them, and therefore any decision rule will make errors when ID and OOD data overlap.} 
\par Many recent OOD works \cite{mohseni2020self}, \cite{zhu2023openmix}, \cite{liu2020energy} follow the idea of adopting auxiliary dataset to regularize the model for improving distinctness between ID and OOD data. These techniques are based on the assumption that the auxiliary datasets represent real OOD data, and, utilizing them as a known priori while training along with ID data can actually aid in generalizing to detect unseen distributions. The benchmark OOD study in using auxiliary OOD data is Outlier Exposure \cite{hendrycks2018deep}. In this paper, authors used a set of outliers that are disjoint from the real OOD test set are used to train the model s to discover signals and learning effective heuristics to detected whether the input belongs to either ID or OOD. While most of these studies follow random sampling of the outliers, other works have considered mining outliers through adversarial training \cite{chen2021atom}, posterior sampling \cite{ming2022poem}, or using leveragin wild mixture data of ID and OOD \cite{katz2022training}.

\subsubsection{Anomaly/Novelty Detection}
\noindent It becomes crucial that a machine learning system be able to distinguish between known and unknown object information during testing, since it is not plausible to train on all potential objects the system is likely to encounter in the real world. In other words, it is crucial that the robust ML systems must have the ability to identify a set of unlabeled instances that significantly differ from the training dataset. Anomaly and novelty detection, often used in tandem throughout the literature, deal with this problem of recognizing anomalous and novel concepts in the system \cite{masana2018metric}. Very subtle difference persists among them, as in the former tries to exclusively find negative samples or pecularities, while the latter focuses on discovering novel concepts that needs to be incorporated into the decision model. Nevertheless, both are concerned over finding the OODness in data where the training samples experiences an abrupt change in concept. It is also to be noted that for any classification system, particularly in data streams, two phenomenon can co-exists: concept evolution, which refers to the emergence of new classes, and concept drift, where the known concept can change over time \cite{faria2016novelty}.

\par In paper \cite{sabokrou2018adversarially}, the authors acknowledged that novelty class in general is either often missing in training, sampled inadequately, or poorly defined, thereby making one-class classifiers a suitable solution for such difficulties. Despite this issue, they proposed an end-to-end architecture specifically for one-class classification, inspired by the success of GANs in training deep models under unsupervised and semi-supervised frameworks. The architecture consisted of two deep networks that are trained together but in opposition, with one network functioning as the novelty detector while the other reinforcing the inlier samples and distorting the outliers. The core idea behind this approach was that the separability between the enhanced inliers and the distorted outliers is substantially greater than when making decisions based on the original samples. Another study \cite{abati2019latent} aimed to tackle the ND challenges, one of which involved recognizing deviations from a typical model of regularity. This task is made difficult by the unpredictable and often undetectable nature of new concepts during training. To address this challenge, the authors created a comprehensive framework that combined a deep AE with a parametric density estimator to learn the underlying probability distribution of latent representations in an autoregressive manner. By optimizing a maximum likelihood objective along with normal sample reconstruction, their approach effectively regularized the task by minimizing the differential entropy of the latent vectors.

\par A prominent technique for self-supervised representation learning is to semantically contrast similar and dissimilar sample pairs \cite{chuang2020debiased}. Considering this, studies such as \cite{chen2020simple},\cite{cho2021masked},\cite{kopuklu2021driver}, \cite{tack2020csi} have utilized contrastive learning (CL) framework for realizing AD task. One of the impressive works by \cite{chen2020simple} exploited task agnostic way of using CL in an AD problem where the agreement between differently augmented views of the same image is maximized while repelling with the others in the same batch. By doing so, this method was able to obtain effective representation of each data sample while robustly clustering each class without the necessity of human supervision or labelling. A following work in  \cite{cho2021masked}, introduced a task-specific variant of CL, termed masked contrastive learning (MCL). Specifically, they unveiled an inference method called self-ensemble inference, designed to enhance performance by exploiting the skills acquired through auxiliary self-supervision tasks. The primary insight of their research was that forming dense clusters, without the necessity for fine-tuning yet preserving individual representations leads to the development of more meaningful visual representations. This approach deviated from the traditional 'pre-train then tune' paradigm practiced by \cite{chen2020simple} and led to effective identification of anomalous data. This study outstood among CL based methods because of its idea of  generating a mask that properly adjusted the repelling ratio while taking into account the class labels present in the batch. 
\subsubsection{Continual Learning}
\noindent Despite the fact that human learning has developed to excel in environments that are constantly changing and evolving, current machine learning systems are only able to perform effectively when presented with well-balanced and homogeneous data. When faced with data that is otherwise, these models often struggle and not only experience a significant decline in performance but also exhibit a catastrophic forgetting phenomenon on previously learned tasks \cite{hadsell2020embracing}. A study in \cite{lesort2021understanding} has termed these phenomenon as interferences that are explicitly caused by the changes in the data distribution or in the learning criterion. Continual learning (CoL), also referred to as lifelong learning \cite{fischer2000lifelong} or incremental learning \cite{castro2018end} share the mutual goal of developing ML algorithms that do not stop learning, but instead keep model parameters updated to accumulate knowledge over time \cite{aljundi2019task}. Modern dynamic data sources can be affected by shifts that can happen over time, (\textit{concept drift}) where the property of some or all classes might abruptly change. Therefore this calls for the demand of CoL models that can effectively adapt to concept drift scenarios in any data stream mining tasks.
\par A study in \cite{schwarz2018progress} has set down requirements for CoL such that; a learning method that continually improves should not experience catastrophic forgetting, meaning it should maintain its ability to perform well on previously learned tasks. Additionally, it should be capable of learning new tasks while leveraging knowledge gained from earlier tasks, demonstrating positive forward transfer for faster learning and improved final performance. The method should also be scalable, able to be trained on a large number of tasks. Furthermore, it should allow for positive backward transfer, meaning that learning a new task can lead to immediate improved performance on previous tasks that are similar or relevant. Lastly, the method should be able to learn without requiring task labels and ideally be applicable in the absence of clear task boundaries.
CoL has traditionally navigated data-constrained scenarios within a supervised framework, where batches of labeled samples were sequentially introduced to the network, enabling it to incrementally assimilate new information while retaining previously acquired knowledge. The research in \cite{taufique2022unsupervised} proposed a method for unsupervised CoL associating unsupervised domain adaptation (UDA) and CoL paradigms. This study addressed the challenge posed by a gradually evolving target domain, segmented into multiple sequential batches, necessitating the model to continuously adapt to the progressively changing data stream without supervision. To address this challenge, they introduced a source-free approach utilizing episodic memory replay coupled with buffer management. Furthermore, a contrastive loss component was integrated to enhance the alignment between the buffer samples and the ongoing flow of batches, aiming to refine the model's adaptability and retention capabilities in the face of evolving datasets.

\par A CoL system is expected to maintain both plasticity, the acquisition of new knowledge, and stability, the preservation of old knowledge. Catastrophic forgetting represents a failure in stability, where new experiences overshadow previous ones. The authors of \cite{rolnick2019experience} utilized replay of past experiences motivated from the approach in neuroscience to mitigate such forgetting. In this work, the authors introduced, a replay-based method designed to significantly diminish catastrophic forgetting within multi-task reinforcement learning environments. The proposed method incorporated off-policy learning and behavioral cloning from replay to strengthen stability, while also employing on-policy learning to ensure the maintenance of plasticity. The paper demonstrated significant performance in mitigating forgetting while referring that the method to be extremely less sophisticated with no requirements of knowledge of tasks being learned. Interestingly, a study was proposed in \cite{caccia2021new} with counterarguments that experience replay leads to significant overlap between the representations of newly added and previous classes, resulting in highly disruptive parameter updates. This study proposed insights to reduce the abrupt change in data representations that occurs when unobserved classes emerge in the data stream. Based on empirical analysis, a new method was proposed to address this problem by protecting the learned representations from drastic adaptations required to accommodate new classes. They showed that using an asymmetric update rule, which encourages new classes to adapt to older ones, is more effective, particularly at task boundaries where significant forgetting typically occurs.

\begin{table*}[htbp]
\centering
\small
\setlength{\tabcolsep}{5pt}
\renewcommand{\arraystretch}{1.35}
\rowcolors{2}{gray!5}{white}
\caption{List of Popular Semantic Shift Mitigation Papers, including their Core Strategy, Best use case, and Limitations.}
\label{tab:sem_proscons}

\resizebox{\textwidth}{!}{%
\begin{tabular}{>{\bfseries}l l p{5.5cm} p{5.5cm} p{5.5cm}}
\rowcolor{gray!20}
\textbf{Type} & \textbf{Reference (Year)} & \textbf{Core Strategy} & \textbf{Best use} & \textbf{Limitation} \\
\midrule

& DeepSAD \cite{ruff2019deep} (2019) & Learns compact representation for normal data via semi-supervised loss & ND with few labeled anomalies & Struggles with complex or overlapping classes \\
& PReNet \cite{pang2023deep} (2023) & Progressive refinement of prediction via contrastive pretext tasks & Fine-grained AD in semantic space & Requires careful task design and tuning \\
& SVM \cite{li2003improving} (2003) & Maximizes margin between classes for robust separation & Classical baseline for AD & Struggles with high-dimensional or nonlinear data \\
& REPEN \cite{pang2018learning} (2018) & Learns distance-aware embeddings for anomaly ranking & Unsupervised outlier detection in high dimensions & Requires sampling strategy; less effective on structured semantics \\
& XGBOD \cite{zhao2018xgbod} (2018) & Boosted ensemble of unsupervised detectors and selected features & Hybrid AD with tabular data & Requires good base detectors; less generalizable \\

\midrule
& OpenMax \cite{bendale2016towards} (2016) & Calibrates softmax scores using extreme value theory & OSR with known–unknown separation & Assumes well-structured class distributions \\
& G-OpenMax \cite{ge2017generative} (2017) & Enhances OpenMax by generating unknowns with GAN & Detecting unknowns in controlled visual domains & Requires high-quality generative models \\
& PROSPER \cite{zhou2021learning} (2021) & Introduces prototype-based unknown classifiers with soft likelihood & Large-scale OSR & Sensitive to prototype quality and feature overlap \\
OSR
& C2AE \cite{oza2019c2ae} (2021) & Combines class-conditioned autoencoders with discriminative embedding & Open-set detection with reconstruction cues & Struggles with visually similar unknown classes \\
& ARPL \cite{chen2021adversarial} (2021) & Learns reciprocal points and margin-based embedding with adversarial training & Open-set robustness with semantic structure & Sensitive to margin settings; adversarial overhead \\
& CSSR \cite{huangClassSpecificSemanticReconstruction2022} (2022) & Reconstructs semantic features via class-specific decoders & Open-set recognition with class-aware reconstruction & Requires reliable class prototypes; complex training \\

\midrule
& PixMix \cite{hendrycks2022pixmix} (2022) & Augments data by mixing with unrelated images & OOD detection under severe distribution shifts & May hurt ID performance; lacks semantic control \\
& LogitNorm \cite{wei2022mitigating} (2022) & Normalizes logits to calibrate softmax scores & OOD detection with pretrained classifiers & May degrade accuracy if misconfigured\\
& OE \cite{hendrycks2018deep} (2018) & Trains model with auxiliary outlier data & Improves OOD detection during training & Depends on quality and diversity of outliers \\
OOD-D
& MCD \cite{zou2024mcd} (2024) & Uses ensemble of classifiers with divergence-based OOD scoring & OOD detection with uncertainty estimation & Computationally heavier; sensitive to ensemble diversity \\
& RotPred \cite{hendrycks2019using} (2019) & Uses self-supervised rotation prediction as an auxiliary task & OOD detection with limited labeled data & Less effective on non-visual or abstract inputs \\
& MixOE \cite{zhang2023mixture} (2023) & Mixes ID and OOD samples for contrastive supervision& Fine-grained OOD detection during training & Requires labeled OOD or curated outliers \\
& AugMix \cite{hendrycks2019augmix} (2019) & Applies diverse and stochastic augmentations with consistency loss & Robust OOD detection and improved generalization & May not capture semantic anomalies \\

\midrule
& ER \cite{rolnick2019experience} (2019) & Stores and replays past samples during training & Simple and effective continual learning & Memory overhead; prone to sampling bias \\
& iCaRL \cite{rebuffi2017icarl} (2017) & Combines rehearsal with nearest-mean-of-exemplars classification & Class-incremental learning & Requires exemplar storage; suffers from imbalance \\
& CVT \cite{wang2022online} (2022) & Online distillation with continual vision transformer adaptation & Continual learning in vision tasks & Requires careful update strategy; compute-heavy \\
CoL
& SCoMMER \cite{sarfraz2023sparse} (2023) & Sparse memory retrieval with modular experts & Lifelong learning with minimal forgetting & Complex memory management; tuning expert modules \\
& DualNet \cite{pham2021dualnet} (2021) & Maintains plastic and stable branches for dual-memory learning & Stability–plasticity trade-off in continual learning & Increased model size and training complexity \\
& BiMeCO \cite{nie2023bilateral} (2023) & Bilateral memory consolidation with contrastive objectives & Continual learning with domain shifts & Sensitive to memory balancing and contrastive tuning \\

\bottomrule
\end{tabular}%
}
\end{table*}

\begin{table*}[htbp]
\centering
\small
\setlength{\tabcolsep}{5pt}
\renewcommand{\arraystretch}{1.35}
\rowcolors{2}{gray!5}{white}
\caption{List of Applied Research Papers Dealing with Covariate Shift Problem}
\label{tab:concept_applied}

\resizebox{\textwidth}{!}{%
\begin{tabular}{>{\bfseries}l l l p{5.5cm} l l}
\rowcolor{gray!20}

\textbf{Type} & \textbf{Reference (Year)} & \textbf{Applied Area} & \textbf{Core Technology} & \textbf{Dataset Type} & \textbf{Used Metrics} \\
\midrule

& Bashari et al. \cite{bashari2024derandomized} (2024) & Intrusion Detection & Conformal inference, RF, SVMs & Synthetic, Network traffic & Power, FDR, Variance \\
& Zhu et al. \cite{zhu2023large} (2023) & Spam Detection & Margin theory, Multi-class novelty detection, SVM & Remote Sensing & AUC, Error \\
& Liu et al. \cite{Liu_2023_CVPR} (2023) & Industrial Defect & Pyramid deformation, Diversity-based detection & Industrial, Surveillance video & AUC \\
AD/ND
& Xu et al. \cite{xu2023data} (2023) & Cyber Intrusion & SMOTE, Multi-class classification & Network traffic & Accuracy, NMI, F1 \\
& Chen et al. \cite{chen2023mgfn} (2023) & Video Anomaly & Feature amplification, Magnitude contrastive loss & Video data & AUC, AP \\
& Xie et al. \cite{xie2023pushing} (2023) & Industrial Anomaly & Few-shot, Graph representation & Industrial images & AUROC \\

\midrule
& Wang et al. \cite{wang2023robustness} (2023) & LLM & Adversarial + OOD robustness, Zero-shot & Product reviews & Adversarial/OOD robustness \\
& Graham et al. \cite{graham2023denoising} (2023) & Image Recon. & Diffusion models, Denoising Autoencoders & Medical images & AUC \\
& Wu et al. \cite{wu2023energy} (2023) & Node Classification & GNNs, Energy Function & Image data & AUROC, AUPR, FPR, ID-Acc \\
OOD-D
& Zhang et al. \cite{zhang2023mixture} (2023) & Image Classification & Mixup, Outlier Exposure & Image data & Accuracy, TNR95 \\
& Wilson et al. \cite{wilson2023hyperdimensional} (2023) & Image Classif. & Hyperdimensional computing, Gram detectors & Image data & AUROC, FPR95, Detection error, F1 \\
& Song et al. \cite{song2023deeplens} (2023) & Text Understanding & Text clustering, Neuron activation & Text data & User rating \\
& Li et al. \cite{li2023rethinking} (2023) & Image Classification & Vision Transformer, One-class finetuning & Image data & AUROC \\
& Mereau et al. \cite{mereaudetecting} (2024) & Text Classification & Max softmax, Mahalanobis KL, Rank-weighted depth & Movie reviews & AUROC, AUPR \\

\midrule
& Liu et al. \cite{liu2023towards} (2023) & Text Recognition & Label-to-prototype, Zero-shot & Chinese text images & Open-set accuracy \\
& Liu et al. \cite{liu2023learning} (2023) & Image Classification & Gaussian prototypes, Bayesian inference & Image data & AUROC, Precision, Recall \\
& Sun et al. \cite{sun2023hierarchical} (2023) & Image Recognition & Hierarchical Attention, LSTM & Image data & Accuracy, AUROC, OSCR \\
& Yang et al. \cite{yang2023progressive} & Image Classification & Model attribution, Sample augmentation & Facial images & AUC, OSCR \\
OSR
& Zhang et al. \cite{zhang2023learning} (2023) & Action Recognition & Reconstruction, Discriminative features & Video data & Accuracy (open/closed set) \\
& Li et al. \cite{li2023gligen} (2023) & Text-to-Image & Diffusion, Zero-shot gen. & Image-text pair & FID, AP \\
& Li et al. \cite{liu2023grounding} (2023) & Object Detection & Language-guided query, Modality fusion & Image-text pair & Accuracy, AP \\
& Soltani et al. \cite{soltani2023adaptable} (2023) & Intrusion Detection & Deep clustering, SVMs & Network traffic & Accuracy, Misclass. error \\

\midrule
& Smith et al. \cite{smith2023closer} (2023) & Image Classification & Rehearsal-free, Param. regularization & Image data & Accuracy, Forgetting \\
& Villa et al. \cite{villa2023pivot} (2023) & Video Classification & Multi-modal classifier, Prompting & Action data & Accuracy, BWF \\
& Raz. et al. \cite{razdaibiedina2023progressive} & Language Model & Progressive prompting, Embedding reparam. & Online reviews & SuperGLUE, FWT, BWT \\
CoL
& Yuan et al. \cite{yuan2023peer} (2023) & Driving Action & P2P federated learning, IoV & Driver video & Objective, Generalizability \\
& Yang et al. \cite{yang2023continual} (2023) & Image Classification & Bayesian GMMs, Incremental learning & Image data & MCR \\
& Zhu et al. \cite{zhu2023continual} (2023) & Semantic Segment. & MDP, Memory sampling, Graph struct. & Image data & Accuracy, IoU \\

\bottomrule
\end{tabular}%
}
\end{table*}

\section{Closely Related Topics}

\subsection{Runtime Monitoring}
\noindent Critical software systems based on ML, such as autonomous vehicles, may exhibit abnormal behavior suddenly, severely, and unpredictably while in operation \cite{topcu2020assured}. Ensuring the safety of these systems is extremely challenging during the design phase. Runtime monitoring is a technique that focuses on monitoring the safety of operation by following the current input and raising an alarm when the safety might be violated, rather than checking the correctness of all inputs universally \cite{hashemi2023runtime}. Such techniques aim to identify unsafe predictions for a given ML model and discard them before they can lead to any catastrophic repercussions. This is usually accomplished by identifying the inputs that are different from the training data \cite{guerin2023out}, \cite{roy2022runtime}.

\subsection{Open World Recognition}
 Open World Recognition (OWR) \cite{bendale2015towards}, \cite{boult2019learning} posits that newly discovered categories ought to be continuously identified and subsequently incorporated into the recognition process for practical applications. In fact, the system must be capable of recognizing objects and assigning them to existing classes, as well as labeling items as unknowns depending on how these objects are distribution-shifted from the learned data. If there are novel instances unknown to the trained model, then they must be gathered and labeled, for example, by humans. Once there is a sufficient quantity of labeled unknowns for class learning, the system must incrementally learn and expand the multi-class classifier, thus rendering each new class "known" to the system. Open World recognition extends beyond mere robustness to unidentified classes and instead aims to create a scalable system that can adapt and learn in an open world amidst the challenging distribution shift phenomenon.

\subsection{Zero-shot Learning}

Zero-shot Learning (ZSL) \cite{romera2015embarrassingly} refers to a method of training a model to classify objects from unseen classes by leveraging knowledge from seen classes through the use of semantic information. Usually, this information is provided in the form of high-dimensional vectors that encompass the names of both the seen and unseen classes. The technique of ZSL essentially bridges the gap between the two types of classes by utilizing semantic information. This approach to learning can be compared to how a human recognizes a new object by assessing the likelihood of its descriptions aligning with previously acquired knowledge. A primal example of this is recognizing a zebra as a horse with black and white stripes, in regards that one has previously encountered horses. We can find several studies of ZSL in the literature for generalized zero shot learning \cite{min2020domain}, and \cite{wu2020self}, zero shot domain generalization \cite{maniyar2020zero}. We guide our readers attention to comprehensive survey papers in ZSL in \cite{pourpanah2022review}, and \cite{wang2019survey}.
\section{Discussion}
{\subsection{Distribution Shift in Large Language Models (LLMs)}}

{Large Language Models (LLMs) are susceptible to vulnerabilities resulting from distribution shifts. An LLM trained on a specific corpus may exhibit reduced performance when there are alterations in the input language \cite{yang2023out}, domain \cite{zhou2022learning}, or task distribution. Such shifts can significantly impair accuracy and increase perplexity, thereby compromising real-world reliability. In the event of a covariate shift, an LLM may encounter difficulties with unfamiliar vocabulary, styles, or structures, leading to misinterpretations. Empirical studies have demonstrated that even large pre-trained models exhibit a notable decline in performance when evaluated on OOD data. For example, in the WILDS benchmark \cite{koh2021wilds} of real-world shifts, models trained on one domain consistently exhibited substantially lower OOD accuracy compared to ID accuracy.} 
\par {Covariate shifts primarily affect the recall of knowledge and alignment to input, where the model may fail to recognize entities or idioms it has not previously encountered (e.g., a medical term abbreviation), or it may incorrectly parse syntax when faced with code or XML in the input. In generation tasks, covariate shifts can lead to incoherence or irrelevant continuations. Notably, LLMs demonstrate some resilience to mild covariate shifts due to their extensive training data; however, under severe shifts, such as transitioning to a different language or a highly specialized jargon, performance can degrade abruptly. This is particularly evident in zero-shot settings where the model has not been conditioned on that style \cite{wang2022language}. Additionally, concept drift may occur if language usage patterns change such that the model’s learned correlations no longer hold (e.g., a word previously indicating negative sentiment is adopted as slang for something positive). Unlike covariate shifts, concept shifts typically alter the decision boundary or generation mapping and may necessitate relearning the task function, such as through fine-tuning on new examples \cite{desale2025concept}. Given the inevitability of distribution drift in real-world data, a variety of strategies have been developed to maintain or improve LLM performance under shift conditions \cite{yuan2023revisiting}. Such strategies encompass retraining or fine-tuning the model with new data, implementing on-the-fly adjustments such as prompting or retrieval, and assessing uncertainty.}

{\subsection{Practical Applications: Impact of Distribution Shift}
\noindent Distribution shifts can profoundly affect a model's performance in practical applications, leading to substantial decrease in the accuracy of downstream tasks. This decline can have serious implications for the decision-making processes in critical systems such as medical diagnosis, autonomous driving, fault detection, intrusion detection, and adversarial defenses. For example, a medical imaging model trained on data from one hospital may not perform well on scans from another due to one of main reasons including the differences in data acquisition device, patient demographics, or imaging protocols. This phenomenon, known as domain shift, can often lead to misdiagnoses or even serious health consequences of the patient. Likewise, autonomous vehicle models trained in sunny conditions may struggle in rain or snow, as their perception systems rely on visual patterns that change with environmental conditions \cite{filos2020can}.} 
\par {In consumer applications such as recommendation systems and spam detection, distribution shifts can result in user dissatisfaction or exploitation. For instance, changes in user behavior over time, such as trends or seasonal interests, can render previously effective recommendation algorithms obsolete, necessitating constant updates \cite{yang2023generic}. In adversarial contexts such as spam or fraud detection, attackers may deliberately induce concept drift, a type of distribution shift to evade detection by exploiting model vulnerabilities \cite{wang2019drifted}. Even in seemingly stable applications like language models or image classifiers, OOD inputs or subtle distribution shifts can lead to high-confidence but incorrect outputs. A prime example of this can be a chatbot trained on standard internet data may produce inappropriate or biased responses when encountering unfamiliar slang, dialects, or cultural contexts. This is particularly concerning in open-world settings, where inputs may come from unpredictable or evolving sources. Overall, distribution shifts undermine the generalization capabilities of machine learning models and reveal the fragility of systems that are not robust to changes in data distribution. They also complicate model evaluation, as performance metrics on held-out test sets may not accurately reflect real-world reliability.}

\subsection{Challenges}
\noindent The distinction between covariate and semantic shifts, while useful for theoretical delineation, may be overly simplistic when addressing practical ML challenges, where these shifts often occur simultaneously and are intertwined. The current categorization - TL, DA, DG/OOD-G for covariate shifts, and OSR, OOD detection, AD/ND, and CoL for semantic shifts - has undoubtedly revolutionized our understanding and capability to tackle each type of shift. Yet, this segregation does not reflect the complexity of real-world applications, where shifts do not present themselves in isolation. For instance, an autonomous driving system may face varying weather conditions (covariate shift) while also encountering new road signs or alterations (semantic shift). Addressing these shifts independently may not be sufficient or efficient for robust performance. While a holistic approach can enhance the algorithm's practical ability to learn from intricate, multifaceted shifts, improving generalization and robustness across diverse situations. In discussing individual methods, we also put forth recently emerging studies that aim to bridge the gap between these two types of shifts, demonstrating the feasibility and efficacy of comprehensive strategies \cite{bai2023feed}, \cite{yang2023full}, \cite{zhang2023nico++}, \cite{tian2021exploring}. These pioneering works indicate that ML algorithms can be designed to be inherently adaptive, and detecting while pursuing to handle both covariate and semantic shifts under one cohesive framework.

\section{Future Research Directions}
 \noindent Developing ML models that can effectively handle data distribution shifts necessitates coordinated efforts across numerous research areas. In order to drive the development of mechanisms to handle the distribution shifts we bring forward several potential future research directions. 

\subfour{Strong Foundation and Benchmarks: }Since the modelling of a unified framework capable of addressing both covariate and semantic shifts simultaneously is of paramount importance, one prospect is to strengthen the current theoretical foundations, which involves formulating comprehensive definitions, metrics, and benchmarks. 

\subfour{Minimal Trade-off: }It is crucial to accomplish a balanced effectiveness in the outcomes of the unified framework for addressing both shifts. Thus, future research should concentrate on developing well-crafted techniques that can achieve effective adaptation and detection without compromising one for the other in a variety of data shift scenarios.

\subfour{Unified Shift Datasets}: Furthermore, innovation in algorithms that can automatically adapt to different types of shifts \cite{gao2024generalize, shao2024open} is crucial, aimed at enhancing model adaptability and robustness with minimal human intervention. Equally important is the establishment of benchmark datasets and evaluation protocols that reflect real-world scenarios involving combined distribution shifts, facilitating more accurate assessments of model performance. 
 
\subfour{Interdisciplinary Approach}: An interdisciplinary approach, incorporating insights from fields such as causality \cite{lv2022causality}, cognitive science \cite{bickhard1995foundational}, and collaboration with domain experts, can be vital to forge direction in implementing novel frameworks to effectively develop solution to tackle data distribution shift problem.

\section{Conclusion}
\noindent In conclusion, this review paper has highlighted the significant obstacles posed by covariate and semantic shifts in ML and emphasized the methodologies inherent in handling these shifts independently. Also, by advocating for an integrated approach, we propose for a paradigm shift towards developing methodologies that cover the entire spectrum of distribution shifts within a single framework. This paper aims to address the current research scenario to the readers and also spark further investigation and innovation paving the way for more efficient, and effective ML applications robust to data distribution shifts.

\ifCLASSOPTIONcompsoc
  \section*{Acknowledgments}
\else
  \section*{Acknowledgment}
\fi

The authors declare no conflict of interests.

\ifCLASSOPTIONcaptionsoff
  \newpage
\fi

\bibliographystyle{IEEEtran}
\bibliography{refs}

\end{document}